\documentclass[runningheads]{llncs}
\usepackage{graphicx}

\usepackage{tikz}
\usepackage{comment}
\usepackage{amsmath,amssymb} 
\usepackage{color}

\usepackage{booktabs}
\usepackage{floatrow}
\usepackage{multicol}
\usepackage{tabularx}
\usepackage{adjustbox}
\usepackage{multirow}
\usepackage{makecell}
\usepackage{xcolor}
\usepackage{placeins}
\usepackage{caption}
\usepackage{subcaption}

\graphicspath{ {./images/} }

\usepackage[accsupp]{axessibility}  

\usepackage[capitalize]{cleveref}

\newfloatcommand{capbtabbox}{table}[][\FBwidth]

\crefname{section}{Sec.}{Secs.}
\Crefname{section}{Section}{Sections}
\Crefname{table}{Table}{Tables}
\crefname{table}{Tab.}{Tabs.}

\begin{document}
\pagestyle{headings}
\mainmatter
\def\ECCVSubNumber{2709}  

\title{Few-shot Image Generation with  Mixup-based Distance Learning} 

\titlerunning{Few-shot Image Generation with MixDL}
%
\author{Chaerin Kong\inst{1}\and
Jeesoo Kim\inst{2}\and
Donghoon Han\inst{1}\and
Nojun Kwak\inst{1}}
\authorrunning{C. Kong et al.}
%
\institute{Seoul National University \and NAVER WEBTOON AI \\
\email{\{veztylord,dhk1349,nojunk\}@snu.ac.kr} \\
\email{jeesookim@webtoonscorp.com}
}
\maketitle

\begin{abstract}
   Producing diverse and realistic images with generative models such as GANs typically requires large scale training with vast amount of images. GANs trained with limited data can easily memorize few training samples and display undesirable properties like ``stairlike" latent space where interpolation in the latent space yields discontinuous transitions in the output space. 
   In this work, we consider a challenging task of pretraining-free few-shot image synthesis, and seek to train existing generative models with minimal overfitting and mode collapse.
   We propose mixup-based distance regularization on the feature space of both a generator and the counterpart discriminator that encourages the two players to reason not only about the scarce observed data points but the relative distances in the feature space they reside. Qualitative and quantitative evaluation on diverse datasets demonstrates that our method is generally applicable to existing models to enhance both fidelity and diversity under few-shot setting. Codes are available\footnote{\url{https://github.com/reyllama/mixdl}}.

\dots
\end{abstract}

\section{Introduction}

\label{sec:intro}

\begin{figure}[t]
  \includegraphics[width=0.9\linewidth]{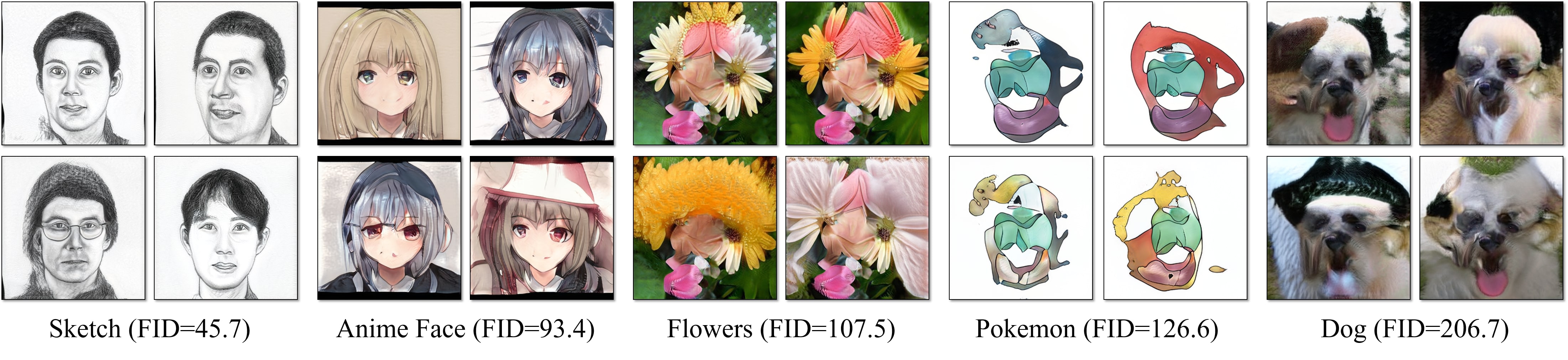}
  \caption{
  \textbf{Cross Domain Correspondence~\cite{ojha2021few} adaptation of FFHQ source generator on various target domains} (10-shot). 
  Finding a semantically similar source domain is crucial for CDC as large domain gap greatly harms the transfer performance. 
  We later show that our method outperforms CDC without any source domain pretraining even on the semantically related domains.
  }
\label{fig:cdc}
\end{figure}

Remarkable features of Generative Adversarial Networks (GANs) such as impressive sample quality and smooth latent interpolation have drawn enormous attention from the community, but what we have enjoyed with little gratitude claim their worth in a data-limited regime. As naive training of GANs with small datasets often fails both in terms of fidelity and diversity, many have proposed novel approaches specifically designed for few-shot image synthesis. Among the most successful are those adapting a pretrained source generator to the target domain \cite{mo2020freeze,ojha2021few,li2020few} and those seeking generalization to unseen categories through feature fusion \cite{gu2021lofgan,hong2020matchinggan}. 
Despite their impressive synthesis quality, these approaches are often critically constrained in practice as they all require semantically related large source domain dataset to pretrain on~\cite{ojha2021few}, as illustrated in \cref{fig:cdc}.
For some domains like abstract art paintings, medical images and cartoon illustrations, it is very difficult to collect thousands of samples, while at the same time, finding an adequate source domain to transfer from is not straightforward either. 
To train GANs from scratch with limited data, several augmentation techniques~\cite{zhao2020differentiable,karras2020training} and model architecture~\cite{liu2020towards} have been proposed. Although these methods have presented promising results on low-shot benchmarks consisting of hundreds to thousands of training images, they fall short for few-shot generation where the dataset is even more constrained (\textit{e.g.,} $n=10$).

\begin{figure}[t]
  \includegraphics[width=\columnwidth]{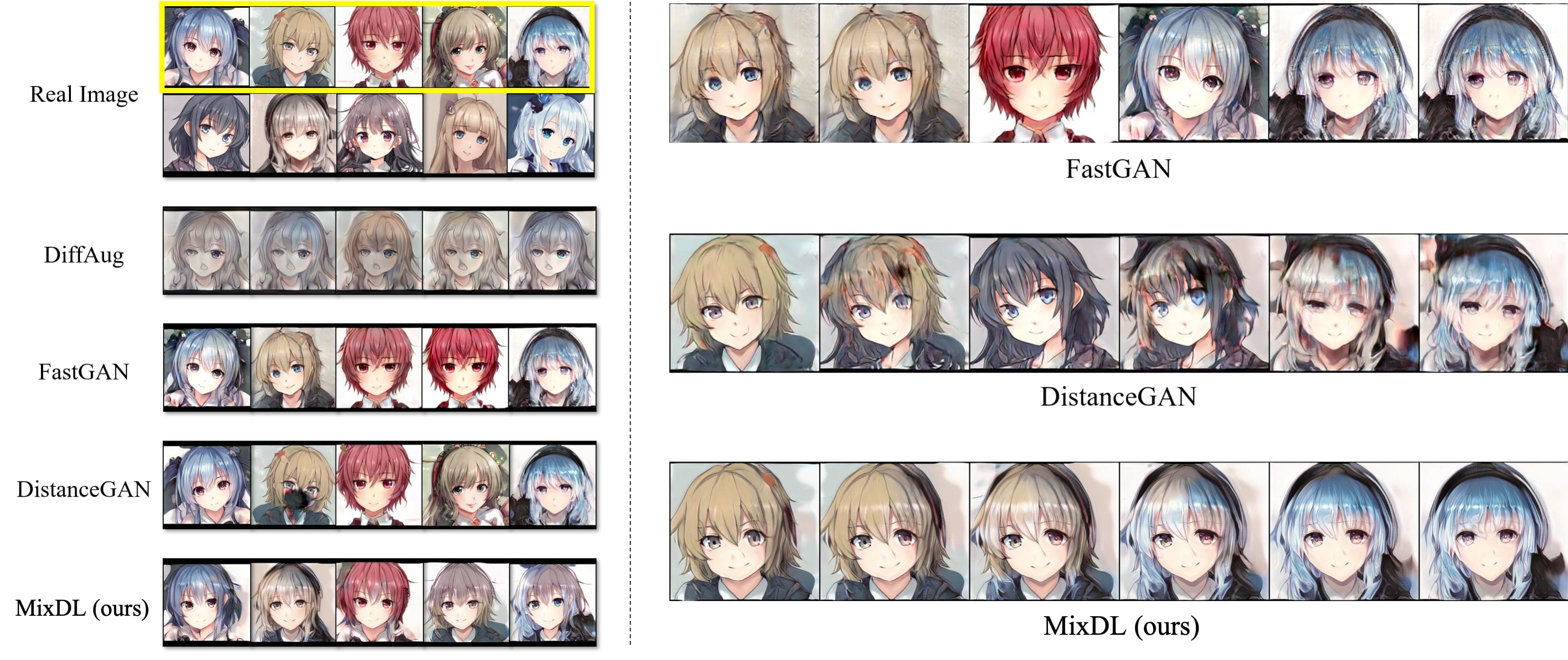}
  \caption{
  Training GANs with as little as 10 real samples typically results in either complete collapse or severe memorization \textit{(left)}. Strongly overfitted generators can only generate a limited set of images, hence displaying \textit{stairlike} latent interpolation \textit{(right)}.
  }
\label{fig:smooth_interp}
\end{figure}

GANs trained with small dataset typically display one of the two behaviors: severe quality degradation~\cite{zhao2020differentiable,karras2020training} or near-perfect memorization~\cite{feng2021gans}, as visible from \cref{fig:smooth_interp} (\textit{left}). 
Hence producing \textit{novel} samples of \textit{reasonable} quality is the ultimate goal of few-shot generative models. 
We note that memorization differs from the classic mode collapse problem, as the former is not just lack of diversity, but the \textit{fundamental inability to generate unseen samples}.

As directly combatting memorization with as little as 10 training samples is extremely difficult if not impossible, we choose to tackle a surrogate problem instead. Our key observation is that strongly overfitted generators are only capable of producing a limited set of samples, resulting in discontinuous transitions in the image space under latent interpolation. We call this \textit{stairlike latent space phenomenon}, which has been pointed out by previous works~\cite{radford2015unsupervised,brock2018large} as an indicator for memorization. \cref{fig:smooth_interp} (\textit{right}) demonstrates that previous methods designed for diversity preservation~\cite{benaim2017one} or low-shot synthesis~\cite{liu2020towards} all display such behavior under few-shot setting ($n=10$). Therefore, instead of pursuing the seemingly insurmountable task of suppressing memorization, we directly target \textit{stairlike latent space problem} and propose effective distance regularizations to explicitly \textit{smooth} the latent space of the generator (G) and the discriminator (D), which we empirically show is equivalent to fighting memorization in effect.



Our high level idea is to maximally exploit the scarce data points by continuously exploring their semantic mixups~\cite{zhang2017mixup}.
The discriminator overfitted to few real samples, however, shows overly confident and abrupt decision boundaries, leaving the generator with no choice but to faithfully replicate them in order to convince the opponent. This results in aforementioned \textit{stairlike latent space} for both G and D, rendering smooth semantic mixups impossible. To tackle this problem, we explore G's latent space with a randomly sampled interpolation coefficient $\mathbf{c}$, enforcing relative semantic distances between samples to follow the mixup ratio. By simultaneously imposing similar regularization on D's feature space, we prohibit the discriminator from embedding images to arbitrary locations for its convenience of memorizing, 
and guide its feature space to be aligned by semantic distances. Our objective is inspired by the formulation of \cite{ojha2021few} that aims to transfer diversity information from source domain to target domain. We tailor it for our single domain setting, where no source domain is available to import diversity from, and show that our method is able to to produce diverse novel samples with convincing quality even with as little as 10 training images.


We further observe that models trained with our regularizations resist mode collapse surprisingly well even with no special augmentation. We believe that our distance regularizations encourage the model to preserve inherent diversity present in early stages throughout the course of training. Resistance to overfitting and mode collapse combined opens up doors for sample diversity under rigorous data constraint, which we demonstrate later with experimental results.

In sum, our contributions can be summarized as:

\begin{itemize}
    \item We propose a two-sided distance regularization that encourages learning of smooth and mode-preserved latent space through controlled latent interpolation.
    \item We introduce a simple framework for few-shot image generation without a large source domain dataset that is compatible with existing architectures and augmentation techniques.
    \item We evaluate our approach on a wide range of datasets and demonstrate its effectiveness in generating diverse samples with convincing quality.
\end{itemize}

\section{Related Works}

\noindent
\textbf{One-shot image generation}
In order to create diverse outcomes from a single image,
SinGAN~\cite{shaham2019singan} leverages the inherent ambiguity present in downsampled image.
Based on SinGAN, ConSinGAN~\cite{hinz2021improved} proposes a technique to control the trade-off between fidelity and diversity.
One-Shot GAN~\cite{sushko2021one} uses a dual-branch discriminator where each head identifies context and layout, respectively.
As one-shot image generation methods focus on exploiting a single image, they are not directly applicable to few-shot image generation tasks where the generator must learn the underlying distribution of a collection of images.

\noindent
\textbf{Low-shot image generation}
Given a limited amount of training data, the discriminator in conventional GAN can easily overfit. To mitigate this problem, 
DiffAugment~\cite{zhao2020differentiable} imposes differentiable data augmentation to both real and fake samples while ADA~\cite{karras2020training} devises non-leaking adaptive discriminator augmentation.
FastGAN~\cite{liu2020towards} suggests a skip-layer excitation module and a self-supervised discriminator, which saves computational cost and stabilizes low-shot training.
GenCo~\cite{cui2021genco} shows impressive results on low-shot image generation task by using multiple discriminators to alleviate overfitting. Despite their promising performances on low-shot benchmarks, these methods often show significant instability under stricter data constraint, namely in \textit{few-shot} setting. 


\noindent
\textbf{Few-shot generation with auxiliary dataset}
Thus far, the \textit{few-shot} image generation task ($n\approx10$) mostly required pretraining on larger dataset with similar semantics~\cite{wang2018transferring,wang2020minegan,zhao2020leveraging,robb2020few} mainly due to its inherent difficulty.
A group of works~\cite{gu2021lofgan,hong2020matchinggan,hong2020f2gan,bartunov2018few} learns transferable generation ability on \textit{seen categories} and seek generalization into \textit{unseen categories} through fusion-based methods.
FreezeD~\cite{mo2020freeze} and EWC~\cite{li2020few} further improves transfer learning framework for GANs.
Meanwhile, CDC~\cite{ojha2021few} computes the similarities between samples within each domain and encourages the corresponding similarity distributions to resemble each other. It aims to directly transfer the structural diversity of the source domain to the target, yielding impressive performance.
In this paper, we modify the formulation of CDC and propose a novel few-shot generation framework that does not require any auxiliary data or separate pretraining step.


\noindent
\textbf{Generative diversity}
Mode collapse has been a long standing obstacle in GAN training. \cite{arjovsky2017wasserstein,mao2017least} introduce divergence metrics that are effective at stabilizing GAN training while \cite{durugkar2016generative,ghosh2018multi} tackle this problem by training multiple networks. Another group of works \cite{liu2019normalized,mao2019mode,tran2018dist,yang2019diversity,benaim2017one} proposes regularization methods to preserve distances in the generated output space. Unlike these works, we consider the few-shot setting where the diversity is restricted mainly due to memorization, and introduce an interpolation-based distance regularization as an effective remedy.

\noindent
\textbf{Latent mixup}
Since \cite{zhang2017mixup}, mixup methods have been actively explored to enforce smooth behaviors in between training samples~\cite{berthelot2019mixmatch,verma2021interpolation,berthelot2019remixmatch}. In generative models, \cite{radford2015unsupervised} emphasizes the importance of smooth latent transition as a counterevidence for memorization, but as state-of-the-art GAN models trained with sufficient data naturally possess such property \cite{karras2020analyzing,brock2018large}, it has been mainly studied with autoencoders. \cite{berthelot2018understanding,oring2020autoencoder} regularize autoencoders to learn smooth latent space while \cite{wertheimer2020augmentation,sainburg2018generative} explore their potential as generative models through interpolation.

\section{Approach}

\begin{figure}[t]
\centering
  \includegraphics[width=0.85\textwidth]{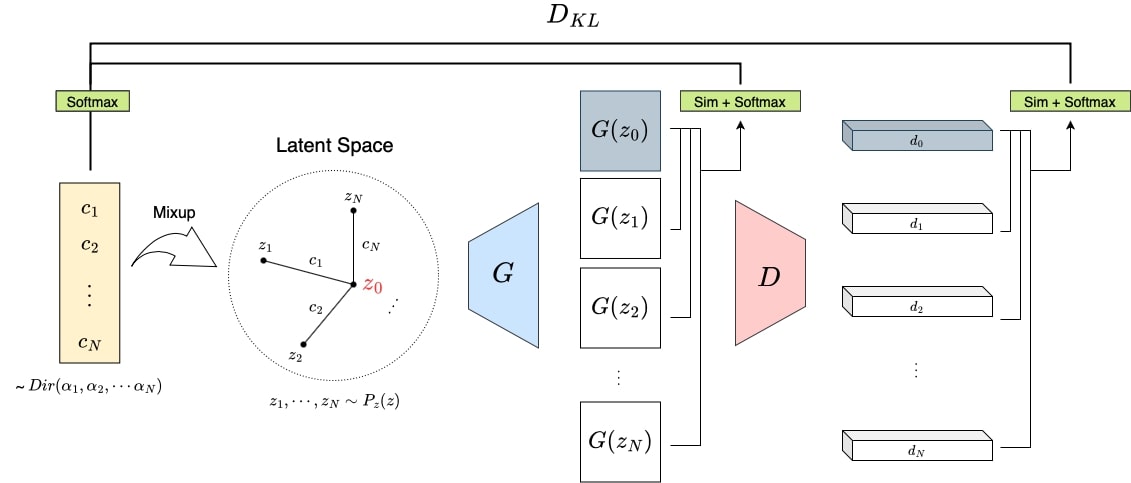}
  \caption{Overview of our \textbf{Mixup-based Distance Learning (MixDL)}. We sample mixup coefficients from a Dirichlet distribution and generate an anchor point $z_0$ through interpolation. Then we enforce pairwise similarities between intermediate generator activations to follow the interpolation coefficients. Similar regularization is imposed on discriminator's penultimate activation, which is linearly projected before similarity calculation. The proposed regularization terms can be added on top of any traditional adversarial framework.}
  \label{fig:overview}
\end{figure}

We consider the situation where only few train examples (e.g., $n=10$) are available with no semantically similar source domain. 
Hence, we would like to train a generative model from scratch, \textit{i.e.,} with no auxiliary dataset or separate pretraining step, using only a handful of images.
Under such challenging constraints, overfitting greatly restricts a model's ability to learn data distribution and produce diverse samples. We identify its byproduct \textit{stairlike latent space} as the core obstacle, as it not only indicates memorizing but also prohibits hallucination through semantic mixup. We observe that both the generator and the discriminator suffer from the problem with insufficient data, evidenced by discontinuous latent interpolation and overly confident decision boundary, respectively. 

To this end, we propose 
mixup-based distance learning (MixDL) framework 
that guides the two players to form soft latent space and leverage it to generate diverse samples. We further discover that our proposed regularizers effectively combat mode collapse, a problem particularly more devastating with a small dataset, by preserving diversity present in early training stages. As our formulation is inspired by \cite{ojha2021few}, we first introduce their approach in \cref{subsection:CDC}, and formally state our methods in \cref{subsection:G} and \cref{subsection:D}. Our final learning framework and the corresponding details can be found in \cref{subsection:Final}.



\subsection{Cross-Domain Correspondence} \label{subsection:CDC}

In CDC~\cite{ojha2021few}, the authors propose to transfer the relationship learned in a source domain to a target domain. They define a probability distribution from pairwise similarities of generated samples in both domains and bind the latter to the former. Formally, they define distributions as
\begin{align}
    p^l &= \text{softmax}(\{\text{sim}(G^l_{s}(z_0), G^l_{s}(z_i))\}_{i=1}^N) \\
    q^l &= \text{softmax}(\{\text{sim}(G^l_{s \rightarrow t}(z_0), G^l_{s \rightarrow t}(z_i))\}_{i=1}^N)
\end{align}
where $G^l$ is the generator activation at the $l^{th}$ layer and $\{z_i\}_0^N$ are latent vectors. Note that $G_{s}$ and $G_{s \rightarrow t}$ correspond to source and target domain generator, respectively, and $p^l$, $q^l$ are $N$-way discrete probability distributions consisting of $N$ pairwise similarities.
Then, along with adversarial objective $\mathcal{L}_{adv}$, they impose a KL-divergence-based regularization of the following form:
\begin{equation}
\mathcal{L}_{dist} = \mathbb{E}_{z \sim p_z(z)}[D_{KL}(q^l||p^l)].
  \label{eqn:loss_dist}
\end{equation}

The benefits of this auxiliary objective are twofold: it prevents distance collapse in the target domain and transfers diversity from the source to target via one-to-one correspondence. However, as visible from \cref{fig:cdc}, the synthesis quality is greatly affected by the semantic distance between source and target. Hence, we propose MixDL, which modifies CDC for pretraining-free few-shot image synthesis and provides consistent performance gains across different benchmarks.

\subsection{Generator Latent Mixup} \label{subsection:G}

In \cite{ojha2021few}, the anchor point $z_0$ could be chosen arbitrarily from the prior distribution $p_z(z)$ since they were transferring the rich structural diversity of the source domain to the target latent space.
As this is no longer applicable in our setting, we propose to resort to diverse \textit{combinations} of given samples. Hence, preserving the modes and learning interpolable latent space are our two main desiderata. To this end, we define our anchor point using Dirichlet distribution as follows:
\begin{equation}
z_0 = \sum_{i=1}^N c_i z_i, \quad \mathbf{c} \sim Dir(\alpha_1, \cdots , \alpha_N)
  \label{eqn:latent_interpolation}
\end{equation}
where $\mathbf{c} \triangleq [c_1, \cdots, c_N]^T$.
Using \cref{eqn:latent_interpolation}, the latent space can be navigated in a quantitatively controlled manner. 
Defining probability distribution of pairwise similarities as in \cite{ojha2021few}, we bind it to the interpolation coefficients $\mathbf{c}$ instead. The proposed distance loss is defined as follows:
\begin{align}
    \mathcal{L}_{dist}^G &= \mathbb{E}_{z \sim p_z(z), \mathbf{c} \sim Dir(\mathbf{\alpha})}[D_{KL}(q^l||p)], \\
    \label{eqn:ours_source} 
    q^l &= \text{softmax}(\{\text{sim}(G^l(z_0), G^l(z_i))\}_{i=1}^N), \\
    p &= \text{softmax}(\{c_i\}_{i=1}^N),
\end{align}
where $Dir(\mathbf{\alpha})$ denotes the Dirichlet distribution with parameters $\alpha=(\alpha_1, \cdots, \alpha_N)$.
This efficiently accomplishes our desiderata. Intuitively, unlike naive generators that gradually converge to few modes, our regularization forces the generated samples to differ from each other by a controlled amount, making mode collapse very difficult. At the same time, we constantly explore our latent space with continuous coefficient vector $\mathbf{c}$, explicitly enforcing smooth latent interpolation. An anchor point similar to \cite{ojha2021few} can be obtained with one-hot coefficients $\mathbf{c}$.

\subsection{Discriminator Feature Space Alignment} \label{subsection:D}

While the generator distance regularization can alleviate mode collapse and stairlike latent space problem surprisingly well, the root cause of constrained diversity still remains unresolved, \textit{i.e.,}\ discriminator overfitting. As long as the discriminator delivers overconfident gradient signals to the generator based on few examples it observes, generator outputs will be strongly pulled towards the small set of observed data. 
To encourage the discriminator to provide smooth signals to the generator based on reasoning about continuous semantic distances rather than simply memorizing the data points, we impose similar regularization on its feature space.
Formally, we define our discriminator $D(x) = (d^{(2)}\circ d^{(1)})(x)$ where $d^{(2)}(x)$ refers to the final FC layer that outputs \{real, fake\}. When a set of generated samples $\{G(z_i)\}_{i=1}^N$ and the interpolated sample $G({z_0})$ is provided to $D$, we construct an $N$-way distribution similar to \cref{eqn:ours_source} as 
%
\begin{equation}
    r = \text{softmax}(\{\text{sim}(proj(d_0^{(1)}), proj(d_i^{(1)}))\}_{i=1}^N)
\end{equation}
where $proj$ refers to a linear projection layer widely used in self-supervised learning literature \cite{chen2020simple,chen2021exploring,grill2020bootstrap} and $d_j^{(1)} \triangleq d^{(1)}(G(z_j))$. Without the linear projector, we found the constraint too rigid 
that it harms overall output quality. We define our distance regularization for the discriminator as

\begin{equation}
    \mathcal{L}_{dist}^D = \mathbb{E}_{z \sim p_z(z), \mathbf{c} \sim Dir(\mathbf{\alpha})} [D_{KL}(r||p)].
\end{equation}

This regularization penalizes the discriminator for storing memorized real samples in arbitrary locations in the feature space and encourages the space to be aligned with relative semantic distances. 
Thus it makes memorization harder while guiding the discriminator to provide smoother and more semantically meaningful signals to the generator.

\subsection{Final Objective} \label{subsection:Final}

\cref{fig:overview} shows an overall concept of our method.
Our final objective takes the form:
\begin{equation}
    \mathcal{L}^G = \mathcal{L}_{adv}^G + \lambda_G\mathcal{L}_{dist}^G
\end{equation}
\begin{equation}
    \mathcal{L}^D = \mathcal{L}_{adv}^D + \lambda_D\mathcal{L}_{dist}^D
\end{equation}
where we generally set $\lambda_G=1000$ and $\lambda_D=1$. 

As our method is largely independent of model architectures, we apply our method to two existing models, StyleGAN2\footnote{https://github.com/rosinality/stylegan2-pytorch}\cite{karras2020analyzing} and FastGAN\cite{liu2020towards}. We keep their objective functions as they are and simply add our regularization terms. For StyleGAN2, we interpolate in $\mathcal{W}$ rather than $\mathcal{Z}$, which has been shown to have better properties such as disentanglement \cite{wang2021high,zhu2020improved,alaluf2021restyle}. Mixup coefficients $\mathbf{c}$ is sampled from a Dirichlet distribution of parameters all equal to one. 
Patch-level discrimination \cite{isola2017image,ojha2021few} is applied for mixup images to encourage our generator to be \textit{creative} while exploring the latent space.

\section{Experiments}

\begin{figure}[t]
\centering
  \includegraphics[width=0.95\textwidth]{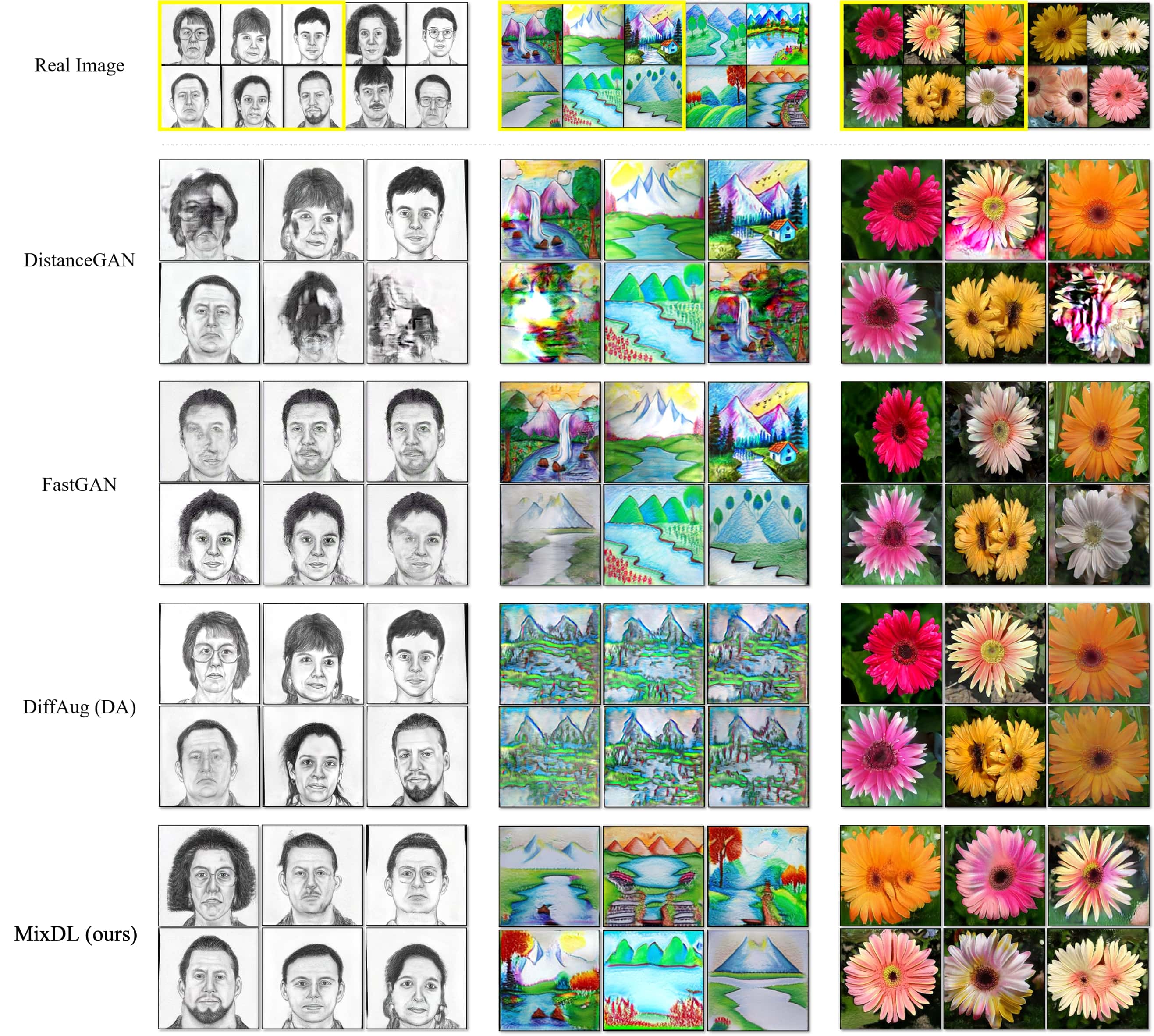}
  \caption{10-shot image generation results. While baseline methods either collapse or \textbf{simply replicate the training samples (yellow box)}, our method actively encourages the generator to explore semantic mixups of given samples, which enables synthesis of various unseen samples.
  }
  \label{fig:generated_samples}
\end{figure}


\noindent
\textbf{Baselines} We mainly apply our method to the state-of-the-art unconditional GAN model, StyleGAN2 \cite{karras2020analyzing}. 
Data augmentation techniques introduced by \cite{zhao2020differentiable} and \cite{karras2020training} show promising performance on low-shot image generation task, so we evaluate them along with ours and refer to them as \textit{DA} and \textit{ADA} respectively. 
We additionally apply our method to FastGAN~\cite{liu2020towards}, which is a light-weight GAN architecture that allows faster convergence with limited data. Although methods designed for alleviating mode collapse~\cite{benaim2017one,liu2019normalized,mao2019mode} are not directly targeted for data-limited setting, we further adopt these as baselines considering the similarity in objective formulation. We implement them on StyleGAN2 for better synthesis quality and fair comparison. Transfer based methods such as EWC~\cite{li2020few} and CDC~\cite{ojha2021few} fundamentally differ from ours as they require a large scale pretraining and thus are not directly comparable. However, we include CDC~\cite{ojha2021few} since our method adjusts it for a more general single domain setting.




\noindent
\textbf{Datasets} For quantitative evaluation, we use Animal-Face Dog \cite{si2011learning}, Oxford-flowers \cite{nilsback2006visual}, FFHQ-babies \cite{karras2019style}, face sketches \cite{wang2008face}, Obama and Grumpy Cat \cite{zhao2020differentiable}, anime face \cite{liu2020towards} and Pokemon (pokemon.com, \cite{liu2020towards}). Aforementioned datasets contain 100 to 8189 samples, so we simulate few-shot setting by randomly sampling 10 images, if not stated otherwise. For qualitative evaluation, we further experiment on paintings of Amedeo Modigliani \cite{yaniv2019face}, landscape drawings \cite{ojha2021few} and web-crawled images of Totoro.  All the images are $256 \times 256$. 
Additional synthesis results and information about datasets can be found in the supplementary.


\begin{figure}[t]
  \includegraphics[width=\linewidth]{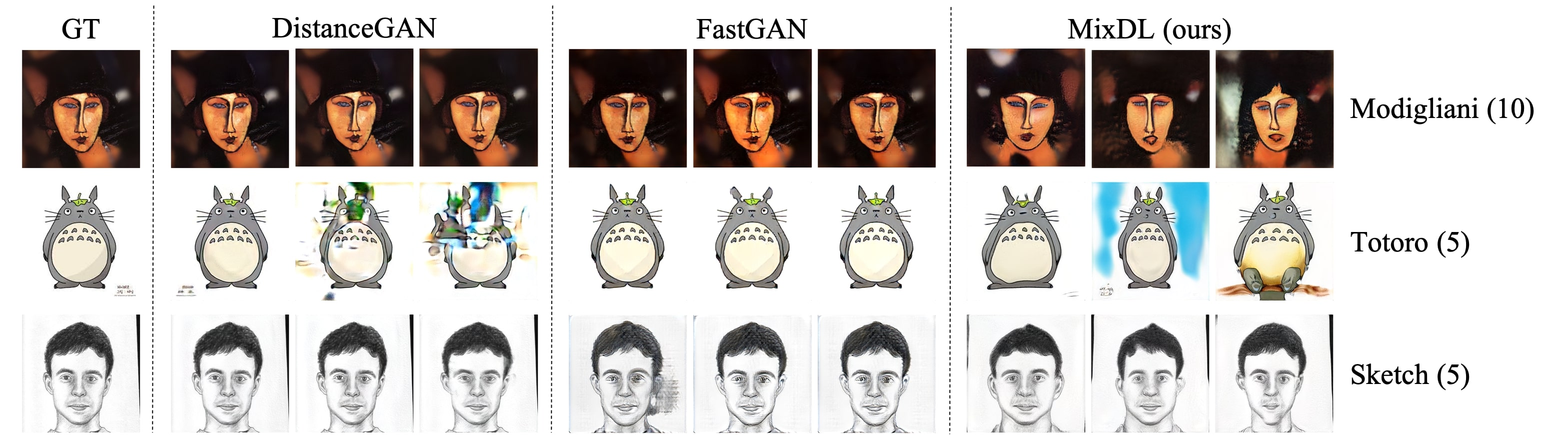}
  \centering
  \caption{
  Uncurated collection of samples sharing the same training image as nearest neighbor. Images from baselines are largely identical, but those produced by ours are all different. Numbers in parentheses indicate the dataset size.
  }
\label{fig:div_under_const}
\end{figure}

\noindent
\textbf{Evaluation Metrics} 
We measure \textit{Fréchet Inception Distance} (FID) \cite{heusel2017gans}, sFID~\cite{nash2021generating} and precision/recall~\cite{zhu2020improved} for datasets containing a sufficient number ($\geq100$) of samples along with pairwise \textit{Learned Perceptual Image Patch Similarity} (LPIPS) \cite{zhang2018unreasonable}. For simulated few-shot tasks, the FID and sFID are computed against the full dataset as in \cite{li2020few,ojha2021few}. 
We further use LPIPS as a distance metric for demonstrating interpolation smoothness and mode preservation.

\subsection{Qualitative Result}



\cref{fig:generated_samples} shows generated samples from 10-shot training. We observe that baseline methods either collapse to few modes or severely overfit to the training data, resulting in inability to generate novel samples. Ours is the only method that produces a variety of convincing samples that are not present in the training set. Our method combines visual attributes such as hairstyle, beard and glasses in a natural way, producing distinctive samples under harsh data constraint.

The difference is more distinguished when we take a closer look. 
In \cref{fig:div_under_const} we display uncurated sets of generated images along with their nearest neighbor real images.
Samples from \textit{DistanceGAN}~\cite{benaim2017one} and \textit{FastGAN}~\cite{liu2020towards} are either defective or largely identical to the corresponding GT, but our method generates unique samples with recognizable visual features.
We believe this is because our distance regularization enforces outputs from different latent vectors to differ from each other, proportionally to the relative distances in the latent space. 

\begin{table}[t]
\caption{\textbf{Quantitative results on 10-shot generation task.} FID and sFID are computed against the full dataset and LPIPS is calculated between generated samples. The best and the second best scores are in bold and underlined. Although CDC$^\dagger$ is not directly comparable as it leverages a pretrained generator (FFHQ), we include it for the relevancy to our method. Clear performance drops are observed with increased domain gap (\textit{e.g.,} FFHQ $\rightarrow$ Dogs). }
\centering
\resizebox{\linewidth}{!}{
\begin{tabular}{l|ccc|ccc|ccc|ccc|ccc}
\Xhline{3\arrayrulewidth}
\multicolumn{1}{c|}{\multirow{2}{*}{Method}} & \multicolumn{3}{c|}{Anime Face} & \multicolumn{3}{c|}{Animal-Face Dog} & \multicolumn{3}{c|}{Oxford Flowers} & \multicolumn{3}{c|}{Face Sketches} & \multicolumn{3}{c}{Pokemon}\\
\cline{2-16}
 & FID $\downarrow$ & sFID $\downarrow$ & LPIPS $\uparrow$ & FID $\downarrow$ & sFID $\downarrow$ & LPIPS $\uparrow$ & FID $\downarrow$ & sFID $\downarrow$ & LPIPS $\uparrow$ & FID $\downarrow$ & sFID $\downarrow$ & LPIPS $\uparrow$ & FID $\downarrow$ & sFID $\downarrow$ & LPIPS $\uparrow$\\
\hline\hline
FastGAN~\cite{liu2020towards} & 123.7 & 127.9 & 0.341 & 103.0 & 117.4 & 0.633 & 182.7 & 111.2 & 0.667 & 76.3 & 81.8 & 0.148 & 123.5 & 105.7 & 0.578 \\
StyleGAN2~\cite{karras2019style} & 166.0 & 111.4 & 0.363 & 177.5 & 127.7 & 0.569 & 177.3 & 143.0 & 0.537 & 94.2 & 84.4 & 0.435 & 257.6 & 136.5 & 0.439 \\
StyleGAN2 + DA~\cite{zhao2020differentiable} & 162.0 & 96.8 & 0.204 & 136.1 & 123.5 & 0.559 & 187.0 & 154.4 & 0.687 & 43.1 & 59.9 & 0.438 & 280.1 & 148.9 & 0.179 \\
StyleGAN2 + ADA~\cite{karras2020training} & 130.2 & 108.0 & 0.288 & 236.5 & 126.2 & 0.636 & 167.8 & 83.5 & 0.719 & 62.8 & 67.3 & 0.399 & 214.3 & 95.5 & 0.496 \\
\hline
FastGAN + Ours & 107.6 & 98.5 & 0.478 & 99.8 & 111.7 & 0.625 & 180.5 & 75.5 & 0.657 & 45.0 & 58.0 & 0.416 & 144.0 & 118.3 & \underline{0.584} \\
StyleGAN2 + Ours & \underline{73.1} & \textbf{92.8} & 0.548 & 96.0 & \underline{99.9} & \underline{0.682} & 136.6 & 67.6 & \underline{0.734} & 39.4 & \textbf{43.3} & \underline{0.479} & \underline{117.0} & \textbf{57.7} & 0.539 \\
StyleGAN2 + DA + Ours & \textbf{70.2} & \underline{94.1} & \underline{0.551} & \underline{96.4} & 107.6 & \underline{0.682} & \underline{129.9} & \underline{66.9} & 0.705 & \textbf{35.6} & 50.1 & 0.471 & \textbf{114.3} & 79.0 & \textbf{0.607} \\
StyleGAN2 + ADA + Ours & 75.0 & 96.5 & \textbf{0.571} & \textbf{94.1} & \textbf{96.6} & \textbf{0.684} & \textbf{127.7} & \textbf{52.5} & \textbf{0.763} & \underline{39.2} & \underline{45.7} & \textbf{0.482} & 155.5 & \underline{65.7} & 0.544 \\
\hline
StyleGAN2 + CDC$^\dagger$~\cite{ojha2021few} & 93.4 & 107.4 & 0.469 & 206.7 & 110.1 & 0.545 & 107.5 & 99.9 & 0.518 & 45.7 & 46.1 & 0.428 & 126.6 & 79.1 & 0.342\\
\Xhline{3\arrayrulewidth}
\end{tabular}
}
\label{tab:quantitative}
\end{table}

\subsection{Quantitative Evaluation}

\cref{tab:quantitative} shows FID, sFID and LPIPS scores for several low-shot generation methods~\cite{zhao2020differentiable,karras2020training,liu2020towards} on 10-shot image generation task. 
We can see that our method consistently outperforms the baselines, often with significant margins. Moreover, our regularizations can be applied concurrently to data augmentations to obtain further performance gains.
Note that while StyleGAN2 armed with advanced data augmentations fails to converge from time to time, our method guarantees stable convergence to a better optimum across all datasets. Surprisingly, ours outperforms CDC~\cite{ojha2021few} on all metrics even when the two domains are closely related, \textit{e.g. anime-face} and \textit{face sketches}. For dissimilar domains like \textit{pokemon}, CDC tends to sacrifice diversity (\textit{i.e.,} LPIPS) for better fidelity, which nevertheless falls short overall. We present training snapshots in the supplementary.

\begin{table}[t]
\caption{Quantitative comparison with diversity preservation methods on 10-shot image generation task. MixDL is equivalent to \textit{StyleGAN2+Ours.}}
\centering
\resizebox{0.9\linewidth}{!}{
\begin{tabular}{c|ccc|ccc|ccc}
\Xhline{3\arrayrulewidth}
\multirow{2}{*}{Method} & \multicolumn{3}{c|}{Anime Face}                & \multicolumn{3}{c|}{Animal-Face Dog}                    & \multicolumn{3}{c}{FFHQ-babies}               \\ \cline{2-10} 
                        & FID $\downarrow$ & sFID $\downarrow$ & LPIPS $\uparrow$ & FID $\downarrow$ & sFID $\downarrow$ & LPIPS $\uparrow$ & FID $\downarrow$ & sFID $\downarrow$ & LPIPS $\uparrow$ \\ \hline \hline
N-Div~\cite{liu2019normalized} & 175.4         & 176.4         & 0.425          & 150.4         & 153.6         & 0.632          & 177.1         & 177.1         & 0.510          \\
MSGAN~\cite{mao2019mode} & 138.6         & 100.5         & 0.536          & 165.7         & 123.0         & 0.630          & 165.4         & 120.1         & 0.569          \\
DistanceGAN~\cite{benaim2017one} & 84.1          & 93.0          & 0.543          & 102.6         & 114.2         & 0.678          & 105.7         & 102.9         & 0.640 \\
MixDL (ours)              & \textbf{73.1} & \textbf{92.8} & \textbf{0.548} & \textbf{96.0} & \textbf{99.9} & \textbf{0.682} & \textbf{83.4} & \textbf{73.9} & \textbf{0.643}          \\
\Xhline{3\arrayrulewidth}
\end{tabular}
}
\label{tab:quant_div}
\end{table}


\begin{figure}[t]
    \begin{floatrow}
\capbtabbox{
\resizebox{1.0\linewidth}{!}{\scriptsize
\begin{tabular}{c|cc|c|cc}
\Xhline{3\arrayrulewidth}

Dataset & Obama & Cat & Flowers & Obama & Cat\\
\hline
Shot & 100 & 100 & 100 & 10 & 10\\
LPIPS & 0.615 & 0.613 & 0.795 & 0.598 & 0.598 \\
\hline
StyleGAN2 & 63.1 & 43.3 & 192.2 & 174.7 & 76.4 \\
+ DA & 46.9 & 27.1 & 91.6 & 66.8 & 45.6 \\
+ Ours & 58.4 & 26.6 & 82.0 & 62.7 & 41.1 \\
+ DA + Ours & \textbf{45.4} & \textbf{26.5} & \textbf{64.0} & \textbf{57.9} & \textbf{39.3} \\
\Xhline{3\arrayrulewidth}
\end{tabular}
}}
{\caption{{\small FID compariosn on low-shot benchmarks. LPIPS measures in-domain diversity.}}
\label{tab:lowshot-benchmark}
}
\capbtabbox{%
\resizebox{0.89\linewidth}{!}{\scriptsize
\begin{tabular}{c|cc|cc}
\Xhline{3\arrayrulewidth}
\multirow{2}{*}{Method} & \multicolumn{2}{c|}{Obama} & \multicolumn{2}{c}{Cat} \\ \cline{2-5} 
                        & Prec.     & Rec.     & Prec.       & Rec.       \\ \hline
StyleGAN2               & 0.47          & 0.07       & 0.15            & 0.12         \\
+MixDL                    & \textbf{0.52}          & \textbf{0.32}       & \textbf{0.86}            & \textbf{0.50}         \\ \hline
FastGAN                 & 0.90          & 0.36       & 0.90            & 0.43         \\
+MixDL                    & \textbf{0.91}          & \textbf{0.47}       & \textbf{0.91}            & \textbf{0.50}         \\
\Xhline{3\arrayrulewidth}
\end{tabular} }}
{\caption{Precision and recall metrics on 100-shot benchmarks.}
\label{tab:prec-recall}
}
\end{floatrow}
\end{figure}

Additional quantitative comparison with diversity preserving methods is displayed in \cref{tab:quant_div}. Although these methods have some similarities with ours, especially MixDL-G, we can observe steady improvements with MixDL. As the baselines are simply designed to minimize mode collapse, we believe they are relatively prone to memorization, which is a far devastating issue in few-shot setting.

While pretraining-free 10-shot image synthesis task has not been studied much, several works \cite{liu2020towards,zhao2020differentiable} have previously explored generative modeling with as little as 100 samples. 
We present quantitative evaluations on popular low-shot benchmarks in Table \ref{tab:lowshot-benchmark}. We observe that our method consistently improves the baseline, and the margin is larger for more challenging tasks, \textit{i.e.,} dataset with greater diversity or fewer training samples. We discuss experiments on these benchmarks in depth in \cref{sec:discussion}. \cref{tab:prec-recall} shows precision and recall~\cite{kynkaanniemi2019improved} for these benchmarks, where MixDL boosts scores especially in terms of diversity.

\subsection{Ablation Study}



We further evaluate the effects of the proposed regularizations, MixDL-G (generator) and MixDL-D (discriminator), through ablation under different settings. In \cref{tab:method_ablation}, we observe that in general, our regularizations both contribute to better quality and diversity, while in some special cases, only adding MixDL-G 
leads to better FID score. 
We conjecture that aligning discriminator's feature vectors with the interpolation coefficients can impose overly strict constraint for some datasets. 
We nonetheless observe consistent improvements on diversity.

\begin{figure}[t]
\centering
\begin{subfigure}{.5\textwidth}
  \centering
  \includegraphics[width=.95\linewidth]{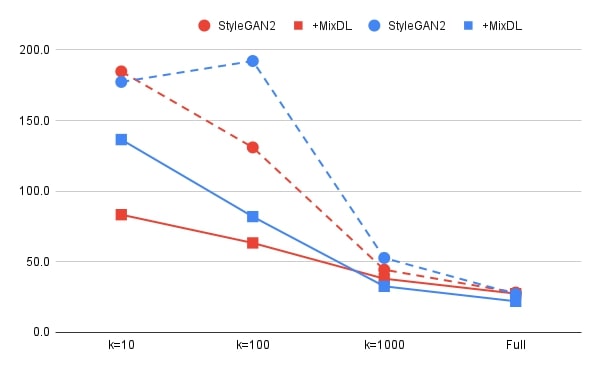}
  \caption{FID scores for different dataset sizes.} 
  \label{fig:shot-fid}
\end{subfigure}%
\begin{subfigure}{.5\textwidth}
  \centering
  \includegraphics[width=.95\linewidth]{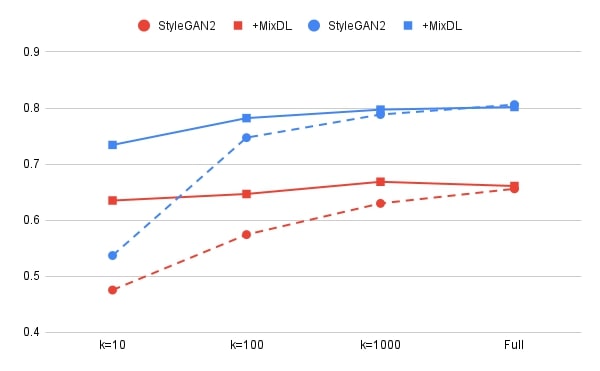}
  \caption{LPIPS for different dataset sizes.}
  \label{fig:shot-lpips}
\end{subfigure}
\caption{Shot ablation results. \textcolor{red}{Red} indicates FFHQ-babies and \textcolor{blue}{blue} represents flowers. Our method consistently improves both metrics with limited data.}
\label{fig:shot-ablation}
\end{figure}

\cref{fig:shot-ablation} shows the evaluation across different subset sizes. Since FFHQ-babies and Oxford-flowers contain more than 2,000 and 8,000 images respectively, we randomly sample subsets of size 10, 100 and 1,000. We can see that the performance of StyleGAN2 steadily improves with more training samples, but it consistently benefits from MixDL. Hence, we believe that with limited data in general, our method can be broadly used to improve model performance. Lastly in \cref{tab:dirichlet_ablation}, the effect of using different Dirichlet concentration parameters and sampling distribution for mixup is illustrated. We find that setting $\alpha=1$ yields the best performance, so we uniformly use this throughout the experiments.

\begin{figure}[t]
\begin{floatrow}
\capbtabbox{
\resizebox{1.0\linewidth}{!}{\scriptsize
\begin{tabular}{cc|cc|cc|cc}
\Xhline{3\arrayrulewidth}
\multicolumn{2}{c|}{MixDL} & \multicolumn{2}{c|}{Dog (10-shot)} & \multicolumn{2}{c|}{Babies (100-shot)} & \multicolumn{2}{c}{Flowers (100-shot)}\\
\hline
$G$ & $D$ & FID $\downarrow$ & LPIPS $\uparrow$ & FID $\downarrow$ & LPIPS $\uparrow$ & FID $\downarrow$ & LPIPS $\uparrow$ \\
\hline
 & & 177.5 & 0.569 & 131.0 & 0.574 & 192.2 & 0.747\\
 & \checkmark & 118.4 & 0.649 & 83.4 & 0.638 & 94.1 & 0.775\\
 \checkmark & & \textbf{95.4} & 0.673 & 71.7 & 0.638 & 84.0 & 0.780\\
 \checkmark & \checkmark & 96.0 & \textbf{0.682} & \textbf{63.4} & \textbf{0.647} & \textbf{82.0} & \textbf{0.782}\\
\Xhline{3\arrayrulewidth}
\end{tabular}
}}
{\caption{{Ablation on MixDL-G and MixDL-D. Two regularizations combined generally yields the best performances.}}
\label{tab:method_ablation}
}
\capbtabbox{%
\resizebox{0.94\linewidth}{!}{\scriptsize
\renewcommand{\arraystretch}{1.6}
\begin{tabular}{c|ccc|c|c}
\Xhline{3\arrayrulewidth}
Distribution & \multicolumn{3}{c|}{Dirichlet} & Gaussian & Uniform \\
\hline
Parameter & $\alpha=0.1$ & $\alpha=1$ & $\alpha=10$ & standard & - \\
\hline
FID ($\downarrow$) & 76.4 & \textbf{73.1} & 80.8 & 76.0 & 74.8\\
LPIPS ($\uparrow$) & 0.536 & \textbf{0.548} & 0.532 & \textbf{0.548} & 0.546\\
\Xhline{3\arrayrulewidth}
\end{tabular}
\renewcommand{\arraystretch}{1}
 }}
{\caption{Mixup coefficient sampling distribution ablation. We adopt $\alpha=1$ for simplicity.}
\label{tab:dirichlet_ablation}
}
\end{floatrow}
\end{figure}

\subsection{Latent Space Smoothness}
\label{sec:latent_smoothness}

Smooth latent space interpolation is an important property of generative models that disproves overfitting and allows synthesis of novel data samples. As our proposed method focuses on diversity through latent smoothing, we quantitatively evaluate this using a variant of Perceptual Path Length (PPL) proposed by \cite{karras2019style}.

\begin{figure}[t]
  \includegraphics[width=\linewidth]{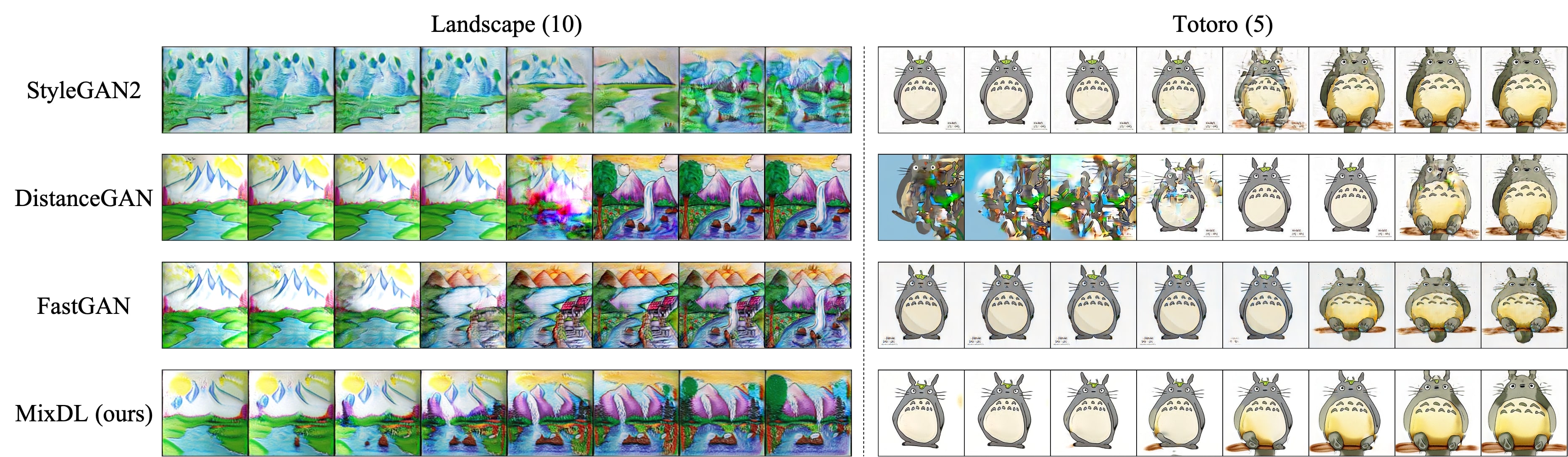}
  \captionsetup{width=\linewidth}
  \caption{\textbf{Latent space interpolation result.} Ours shows smooth transitions with high quality while others show defective or abrupt transitions.}
\label{fig:latent_interp}
\end{figure}

PPL was originally introduced as a measure of latent space disentanglement under the assumption that a more disentangled latent space would show smoother interpolation behavior \cite{karras2019style}. As we wish to directly quantify latent space smoothness, we slightly modify the metric by taking 10 subintervals between any two latent vectors and measure their perceptual distances. \cref{tab:ppl_uniform} reports the subinterval mean, standard deviation, and the mean for the full path (\textit{End}). Note that as PPL is a quadratic measure, the sum of subinterval means can be smaller than the endpoint mean.
All four models show similar endpoint mean, suggesting that the overall total perceptual distance is consistent, while ours displays the lowest PPL standard deviation. As low PPL variance across subintervals is a direct sign of perceptually uniform latent transitions, we can verify the effectiveness of our method in smoothing the latent space. Similar insight can be found from \cref{fig:latent_interp} where the baselines display \textit{stairlike} latent transition while ours shows smooth semantic interpolation. More details on PPL computation can be found in the supplementary materials.





\subsection{Preserving Diversity}


\begin{table}[t]
\caption{\textbf{Perceptual Path Length uniformity.} We generate 5000 latent interpolation paths and subdivide each into 10 subintervals to compute perceptual distances. Standard deviation (std) is computed across the subintervals, indicating perceptual uniformity of latent transition.}
\centering
\resizebox{0.8\linewidth}{!}{
\begin{tabular}{>{\centering}p{0.2\linewidth}|>{\centering}p{0.1\linewidth}>{\centering}p{0.1\linewidth}>{\centering}p{0.1\linewidth}|>{\centering}p{0.1\linewidth}>{\centering}p{0.1\linewidth}>{\centering\arraybackslash}p{0.1\linewidth}}
\Xhline{3\arrayrulewidth}
\multicolumn{1}{c|}{Dataset} & \multicolumn{3}{c|}{Landscape} & \multicolumn{3}{c}{Totoro}\\
\hline
\multicolumn{1}{c|}{Metric} & Mean & Std. & End & Mean & Std. & End \\
\hline
StyleGAN2 & 21.91 & 12.66 & 60.90 & 16.43 & 15.39 & 56.53 \\
DistanceGAN & 23.07 & 21.53 & 70.71 & 16.76 & 14.82 & 61.50 \\
FastGAN & 15.49 & 15.00 & 67.75 & 10.03 & 12.14 & 54.16 \\
MixDL & 12.82 & \textbf{4.19} & 64.28 & 11.75 & \textbf{6.44} & 56.83 \\
\hline
\end{tabular}
}
\label{tab:ppl_uniform}
\end{table}


\begin{figure}[t]
\centering
\begin{subfigure}{.5\textwidth}
  \centering
  \includegraphics[width=.95\linewidth]{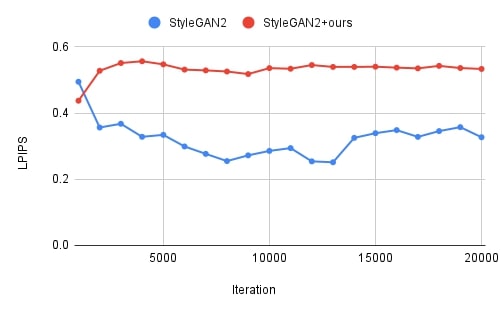}
  \caption{LPIPS in early iterations.} 
  \label{fig:diversity1}
\end{subfigure}%
\begin{subfigure}{.5\textwidth}
  \centering
  \includegraphics[width=.95\linewidth]{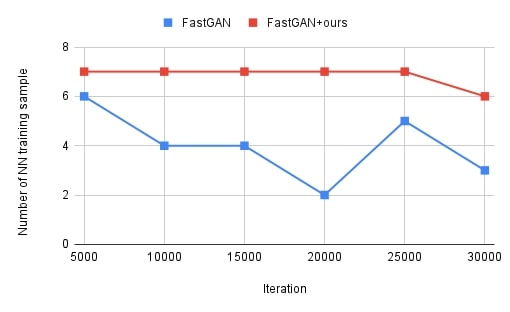}
  \caption{Number of unique NN training samples.}
  \label{fig:diversity2}
\end{subfigure}
\caption{Analysis on sample diversity. (a) shows that our method produces samples with greater diversity. (b) indicates the number of unique training samples that are nearest neighbor to any of the generated samples. We generate 500 samples for the analysis. Since we train our model with 10 samples, the upper bound is 10. Training snapshots are available in the supplementary materials.}
\label{fig:diversity}
\end{figure}

As opposed to \cite{ojha2021few} that preserves diversity in the source domain, our method can be interpreted as preserving the diversity inherently present in the early stages throughout the course of training, by constantly exploring the latent space and enforcing relative similarity/difference between samples. 
To validate our hypothesis, we keep track of pairwise LPIPS of generated samples and the number of \textit{modes} in the early iterations. 
\cref{fig:diversity} shows the result, where the number of \textit{modes} is represented by the number of unique training samples (real images) that are the nearest neighbor to any of the generated images. 
In \cref{fig:diversity1}, we can see that vanilla StyleGAN2 and our method show similar LPIPS in the beginning, but the baseline quickly loses diversity as opposed to ours that maintain relatively high level of diversity throughout the training. \cref{fig:diversity2} delivers similar implication that FastGAN trained with our method better preserves modes, thus diversity, compared to the baseline.

Combined with latent space smoothness explained in \cref{sec:latent_smoothness}, generators equipped with MixDL learn rich mode-preserving latent space with smooth interpolable landscape. This naturally allows generative diversity particularly appreciated under the constraint of extremely limited data.

\section{Discussion}

\label{sec:discussion}  


The trade-off between fidelity and diversity in GANs has been noted by many~\cite{brock2018large,karras2019style}. Truncation trick, a technique widely used in generative models, essentially denotes that diversity can be traded for fidelity. In few-shot generation task, it is very straightforward to obtain near-perfect fidelity at the expense of diversity as one can simply overfit the model, while generating diverse \textit{unseen} data points is very challenging. This implies that with only a handful of data, the diversity should be credited no less than the fidelity.

However, we believe that the widely used low-shot benchmarks, e.g., 100-shot Obama and Grumpy Cat, inherently favor faithful reconstruction over audacious exploration.
The main limitations we find in these datasets are twofold: (i) the intra-diversity is too limited as they contain photos of a single person or object, evidenced by low LPIPS in \cref{tab:lowshot-benchmark} and (ii) FID is computed based on the 100 samples that were used for training.
We acknowledge that (ii) is a common practice in generative models, but the problem with these benchmarks is that the number of samples is too limited, making it possible for some models to simply \textit{memorize} a large portion of them. 
These two combined results in benchmarks that allow relatively easy replication and reward it generously at the same time. 
In other words, we believe that a model's capacity to explore continuous image manifold and \textit{be creative} can potentially backfire in these benchmarks. 

To address these limitations, in \cref{tab:lowshot-benchmark} we extend the benchmark with three additional datasets: 100-shot Oxford-flowers, 10-shot Obama and Grumpy Cat. The first one challenges the model with greater diversity while the last two evaluate its capacity to learn distribution in a generalizable manner, as the FID is still computed against the full 100 images.
As our method mainly aims for modeling diversity, we observe marginal performance gains in the traditional benchmarks. 
However on the extended benchmarks, our proposed method shows significant contributions, confirming that it excels at learning diversity even under challenging situations.

\section{Conclusion}

We propose MixDL, a set of distance regularizations that can be directly added on top of existing models for few-shot image generation. Unlike previous works, MixDL enables high-quality synthesis of novel images with as few as 5 to 10 training samples, even without any source domain pretraining. Thorough evaluations on diverse benchmarks consistently demonstrate the effectiveness of our framework. We hope our work facilitates future research on data efficient generative modeling, which we believe has great upside in both academics and practical applications.


\clearpage
%
%

\renewcommand\thesection{\Alph{section}}
\newcommand{\beginsupplement}{%
        \setcounter{table}{0}
        \renewcommand{\thetable}{S\arabic{table}}%
        \setcounter{figure}{0}
        \renewcommand{\thefigure}{S\arabic{figure}}%
     }

\title{Supplementary Materials} 

\titlerunning{Few-shot Image Generation with MixDL}



\author{}
\institute{}

\maketitle
\beginsupplement


\newcommand\nj[1]{\textcolor{red}{#1}}


\section{Implementation Details}

\noindent
\textbf{StyleGAN2} We adopt the standard StyleGAN2 architecture\footnote{https://github.com/rosinality/stylegan2-pytorch} for $256\times256$ resolution images, with 8 fully connected layers in the mapping network. We keep the hyperparameters such as the learning rate, regularization weights and frequency, untouched, and only add our proposed MixDL. 

\noindent
\textbf{DiffAug} We essentially follow the official configuration\footnote{https://github.com/mit-han-lab/data-efficient-gans} for \textit{low-shot} generation, including the two-layer mapping network and three data augmentation methods. We have also tried with a standard 8 FC layer mapping network and observed significant drops in the overall performance as shown in \cref{tab:diffaug-lowshot}. 

\begin{table}[h]
\caption{FID for DiffAug with varying number of FC layers}
\centering
\begin{tabular}{c|cc}
\Xhline{3\arrayrulewidth}
FC layers & Obama (100-shot) & Grumpy Cat (100-shot)\\
\hline
2 & 46.87 & 26.52 \\
8 & 71.13 & 38.42 \\
\hline
\end{tabular}
\label{tab:diffaug-lowshot}
\end{table}

\noindent
\textbf{FastGAN} We use the official FastGAN implementation\footnote{https://github.com/odegeasslbc/FastGAN-pytorch} for $256\times256$ images. As FastGAN doesn't have a separate mapping network, we interpolate in $\mathcal{Z}$ space. 


\noindent
\textbf{Diversity Preservation Methods} Baselines such as \textit{Normalized Diversification (N-Div)} [28], \textit{Mode Seeking GAN (MSGAN)} [29] and \textit{DistanceGAN} [4] propose distance preserving objective to combat mode collapse. We train these models with StyleGAN2 architecture for better synthesis quality and fair comparison.

\noindent
\textbf{MixDL} For MixDL, we alternate between the normal adversarial training step and the interpolation/regularization step. In the former we go through normal image-level discrimination and in the latter, we apply patch-level discrimination on the mixup samples and compute losses for MixDL-G and MixDL-D. For patch discrimination, we largely adopt the implementation of Cross-domain Correspondence (CDC)\footnote{https://github.com/utkarshojha/few-shot-gan-adaptation}. Our linear projection layer for the discriminator operates on 512 dimension.

\noindent
\textbf{Percpetual Path Length} For PPL computation, we mainly follow the implementation in StyleGAN2. The difference is that we subdivide a latent interpolation path into 10 subintervals and compute the perceptual distance for each line segment. Since the original PPL computation divides the perceptual distance by the squared step size, we divide each subinterval length by $0.1^2$. For clear demonstration, we divide the endpoint mean by $0.1^2$ as well. Note that the overall procedure is equivalent to calculating LPIPS multiplied by the factor of 100. The standard deviation is computed across the subintervals, and averaged for the interpolation paths.

\noindent
\textbf{Number of Modes} We generate 500 samples and compute their perceptual distances to the 10 training samples. We record the index for the real sample with the smallest perceptual distance and report the unique count. It is visually apparent from \cref{fig:snapshots1} that our method boosts mode diversity.


\section{Datasets}

We present the datasets used in our work along with their size.

\begin{table}[h]
\caption{Number of shots used in each dataset. \textbf{Names of datasets} are presented in the first and third rows and their corresponding \textbf{number of shots} used in this paper are described in the second and fourth rows.}
\centering
\resizebox{0.95\linewidth}{!}{
\renewcommand{\arraystretch}{1.3}
\begin{tabular}{c|c|c|c|c|c}
\Xhline{3\arrayrulewidth}
\makecell{Animal-Face\\Dog} & \makecell{Oxford Flowers} & \makecell{FFHQ Babies} & \makecell{Sketches} & \makecell{Obama} & \makecell{Grumpy\\Cat} \\
\hline
 10 & 10, 100, 1000, 8192 & 10, 100, 1000, 2479 & 5, 10 & 10, 100 & 10, 100 \\
\hline
\makecell{Pokemon} & \makecell{Amedeo Modigliani} & \makecell{Anime Face} & \makecell{Landscape} & \makecell{Totoro} & \makecell{} \\
\hline
10 & 10 & 10 & 10 & 5 &  \\
\Xhline{3\arrayrulewidth}
\end{tabular}
\renewcommand{\arraystretch}{1.0}
}
\vspace{-2mm}
\label{tab:dataset}
\end{table}

\section{Additional Evaluations with CDC [34]}

We provide evaluation results for CDC [34] on two popular low shot benchmarks, Obama and Cat (\cref{tab:additional}). To simulate few-shot setting, we randomly sample 10 images from each dataset.. Since CDC is pretrained on FFHQ, the domain gap is relatively small, especially for Obama dataset. Nevertheless, we observe superior performances with MixDL.

\begin{table}[h]
\caption{FID, precision and recall are computed against the full dataset (with 100 images) while LPIPS is computed among the generated samples.}
\centering
\resizebox{0.95\linewidth}{!}{
\begin{tabular}{c|cccc|cccc}
\Xhline{3\arrayrulewidth}
     & \multicolumn{4}{c|}{Obama (10-shot)}                           & \multicolumn{4}{c}{Cat (10-shot)}                              \\ \hline
Model & FID($\downarrow$)           & LPIPS($\uparrow$)          & Prec.($\uparrow$)        & Rec.($\uparrow$)         & FID($\downarrow$)          & LPIPS($\uparrow$)          & Prec.($\uparrow$)        & Rec.($\uparrow$)         \\ \hline
CDC   & 75.0          & 0.490          & 0.47          & 0.07          & 45.3          & 0.451          & 0.52          & 0.10          \\
MixDL   & \textbf{62.7} & \textbf{0.601} & \textbf{0.53} & \textbf{0.09} & \textbf{41.1} & \textbf{0.590} & \textbf{0.78} & \textbf{0.11} \\ \hline
\end{tabular}
}
\label{tab:additional}
\end{table}

\section{Additional Baseline Comparisons}

We present quantitative evaluation results with concurrent competitive baselines~\cite{tseng2021regularizing,cui2021genco} in combination to different data augmentations in \cref{tab:add_baselines}. We observe consistent benefits from MixDL.

\begin{table}[h]
\caption{Comparison with additional baselines. MixDL consistently outperforms others even without advanced augmentations.}
\centering
\resizebox{0.95\linewidth}{!}{\scriptsize
\begin{tabular}{l|cc|cc|cc|cc}
\Xhline{3\arrayrulewidth}
\multicolumn{1}{c|}{Dataset} & \multicolumn{2}{c|}{Anime-face}                     & \multicolumn{2}{c|}{Dog}                            & \multicolumn{2}{c|}{Flower}                          & \multicolumn{2}{c}{Baby}                           \\ \hline
\multicolumn{1}{c|}{Metric}  & \multicolumn{1}{c}{FID}           & LPIPS          & \multicolumn{1}{c}{FID}           & LPIPS          & \multicolumn{1}{c}{FID}            & LPIPS          & \multicolumn{1}{c}{FID}           & LPIPS          \\ \hline
LeCam + DA            & \multicolumn{1}{c}{286.7}         & 0.130          & \multicolumn{1}{c}{129.7}         & 0.593          & \multicolumn{1}{c}{189.2}          & 0.688          & \multicolumn{1}{c}{127.7}         & 0.588          \\
GenCo + DA            & \multicolumn{1}{c}{222.4}         & 0.082          & \multicolumn{1}{c}{147.2}         & 0.565          & \multicolumn{1}{c}{186.1}          & 0.702          & \multicolumn{1}{c}{119.3}         & 0.605          \\
MixDL + DA                      & \multicolumn{1}{c}{\textbf{70.2}} & \textbf{0.551} & \multicolumn{1}{c}{\textbf{96.4}} & \textbf{0.682} & \multicolumn{1}{c}{\textbf{129.9}} & \textbf{0.705} & \multicolumn{1}{c}{-}             & -              \\ \hline 
LeCam + ADA           & \multicolumn{1}{c}{111.6}         & 0.405          & \multicolumn{1}{c}{239.0}         & 0.378          & \multicolumn{1}{c}{191.0}          & 0.659          & \multicolumn{1}{c}{178.3}         & 0.451          \\
GenCo + ADA           & \multicolumn{1}{c}{93.7}          & 0.450          & \multicolumn{1}{c}{112.4}         & 0.652          & \multicolumn{1}{c}{194.0}          & 0.673          & \multicolumn{1}{c}{103.8}         & 0.570          \\
MixDL + ADA                     & \multicolumn{1}{c}{\textbf{75.0}} & \textbf{0.571} & \multicolumn{1}{c}{\textbf{94.1}} & \textbf{0.684} & \multicolumn{1}{c}{\textbf{127.7}} & \textbf{0.763} & \multicolumn{1}{c}{-}             & -              \\ \hline 
MixDL (no aug.)                 & \multicolumn{1}{c}{73.1}          & 0.548          & \multicolumn{1}{c}{96.0}          & 0.682          & \multicolumn{1}{c}{136.6}          & 0.734          & \multicolumn{1}{c}{\textbf{83.4}} & \textbf{0.643} \\ \hline
\end{tabular}
}
\label{tab:add_baselines}
\vspace{-0.5mm}
\end{table}

\section{Training Snapshots}

\begin{figure}[t]
\includegraphics[width=\linewidth]{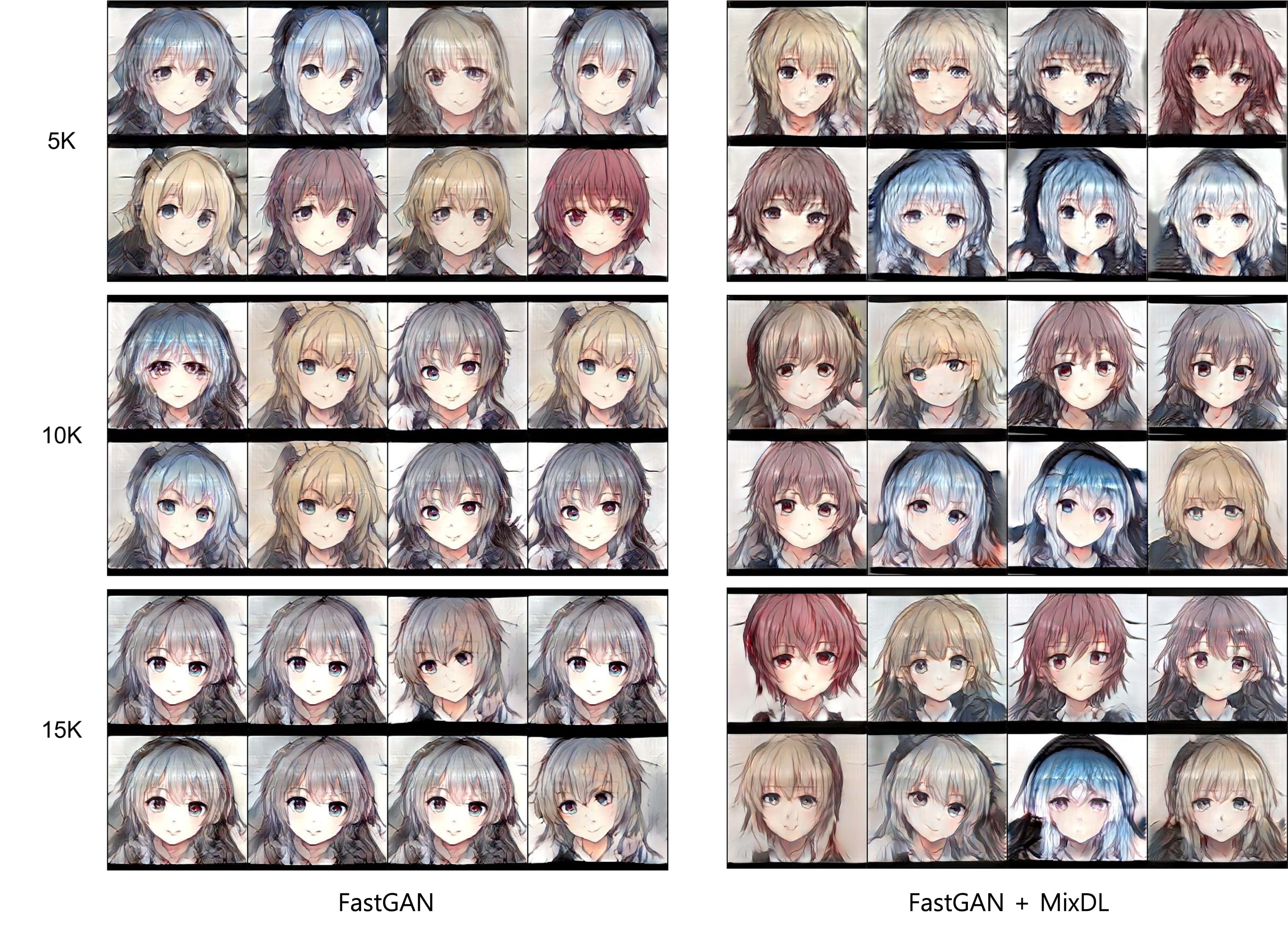}
  \caption{
  Training snapshots for FastGAN and FastGAN+MixDL in early iterations. As opposed to the base FastGAN that rapidly loses diversity, our regularizations help preserve the modes throughout the course of training. Numbers in the left indicate training iterations.
  }
\label{fig:snapshots1}
\end{figure}

\begin{figure*}
\includegraphics[width=0.9\linewidth]{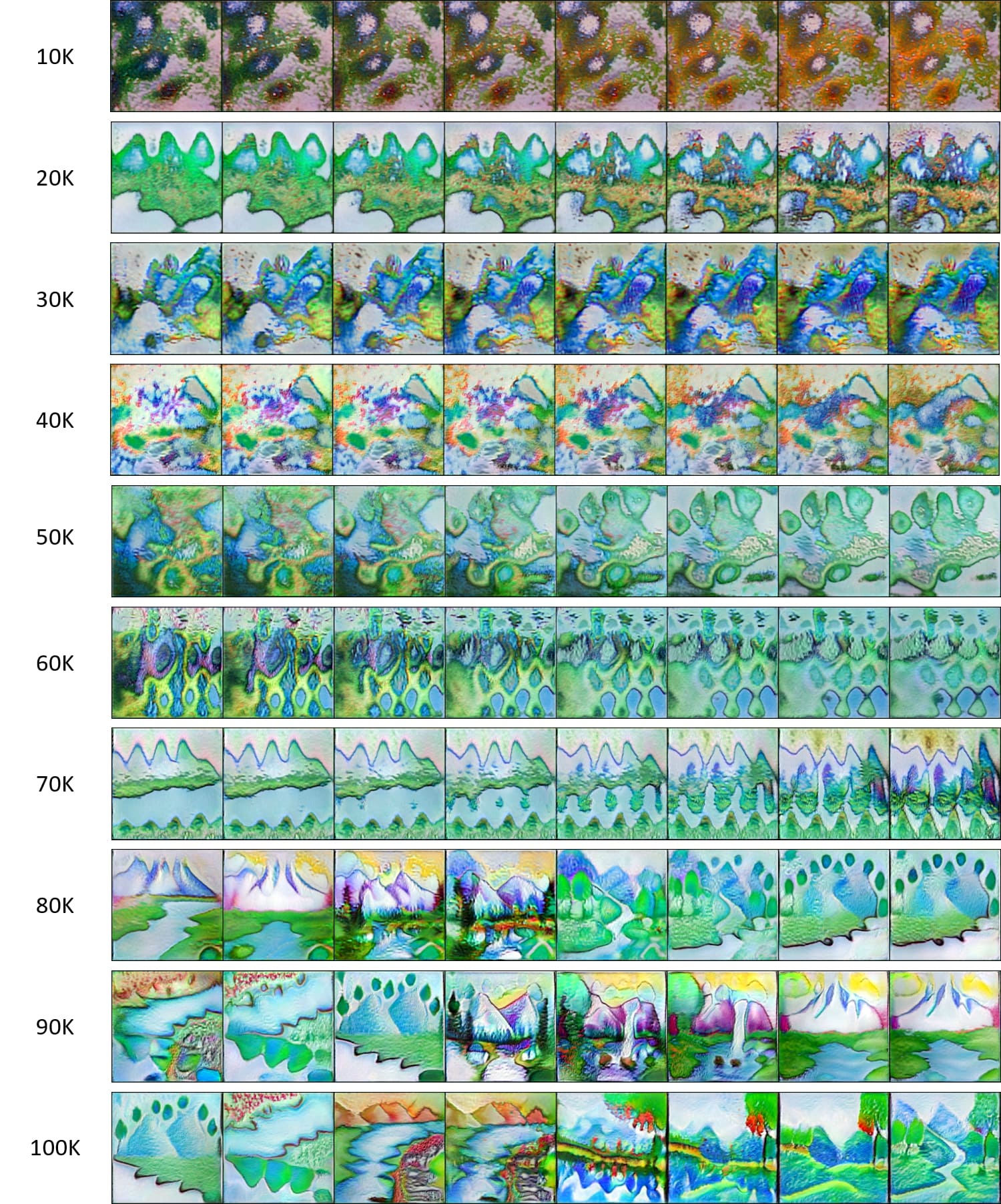}
  \caption{
  Interpolation snapshots for StyleGAN2. Numbers in the left indicate training iterations.
  }
\label{fig:snapshots2}
\end{figure*}

\begin{figure*}
\includegraphics[width=0.9\linewidth]{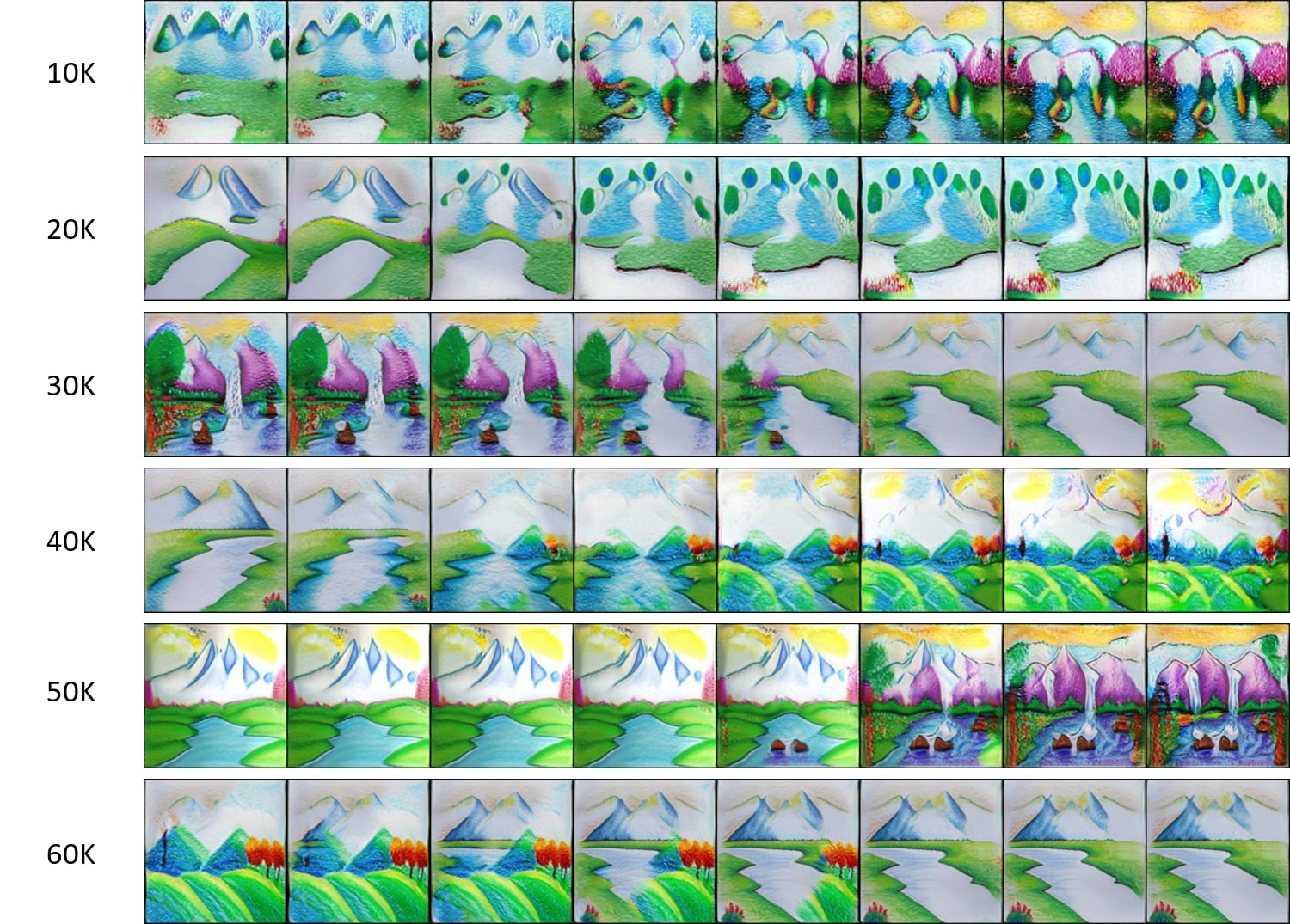}
  \caption{
  Interpolation snapshots for StyleGAN2+MixDL.
  }
\label{fig:snapshots3}
\end{figure*}

We provide training snapshots for FastGAN and StyleGAN2 for visual demonstration of diversity and interpolation smoothness. \cref{fig:snapshots1} clearly shows that as opposed to vanilla FastGAN that rapidly loses diversity and converges to few prototypes, MixDL successfully alleviates this. \cref{fig:snapshots2} displays interpolation snapshots for StyleGAN2. In early training iterations, it does show relatively smooth latent transition, but the sample quality is very unsatisfactory. As the training proceeds, the sample quality improves as the model \textit{overfits}, but consequently the interpolation smoothness is quickly lost. This describes the classic dilemma in few-shot generative modeling. In contrast, \cref{fig:snapshots3} shows that as MixDL is effective at maintaining latent space smoothness, it provides a sweet spot where reasonable sample quality and smooth latent transition coexist. Note that models with MixDL do inevitably overfit in the end, but we can find reasonable stopping point that produces diverse unseen samples with satisfactory visual quality.

\section{Additional Generated Samples}

\begin{figure}
\centering
\includegraphics[width=0.95\linewidth]{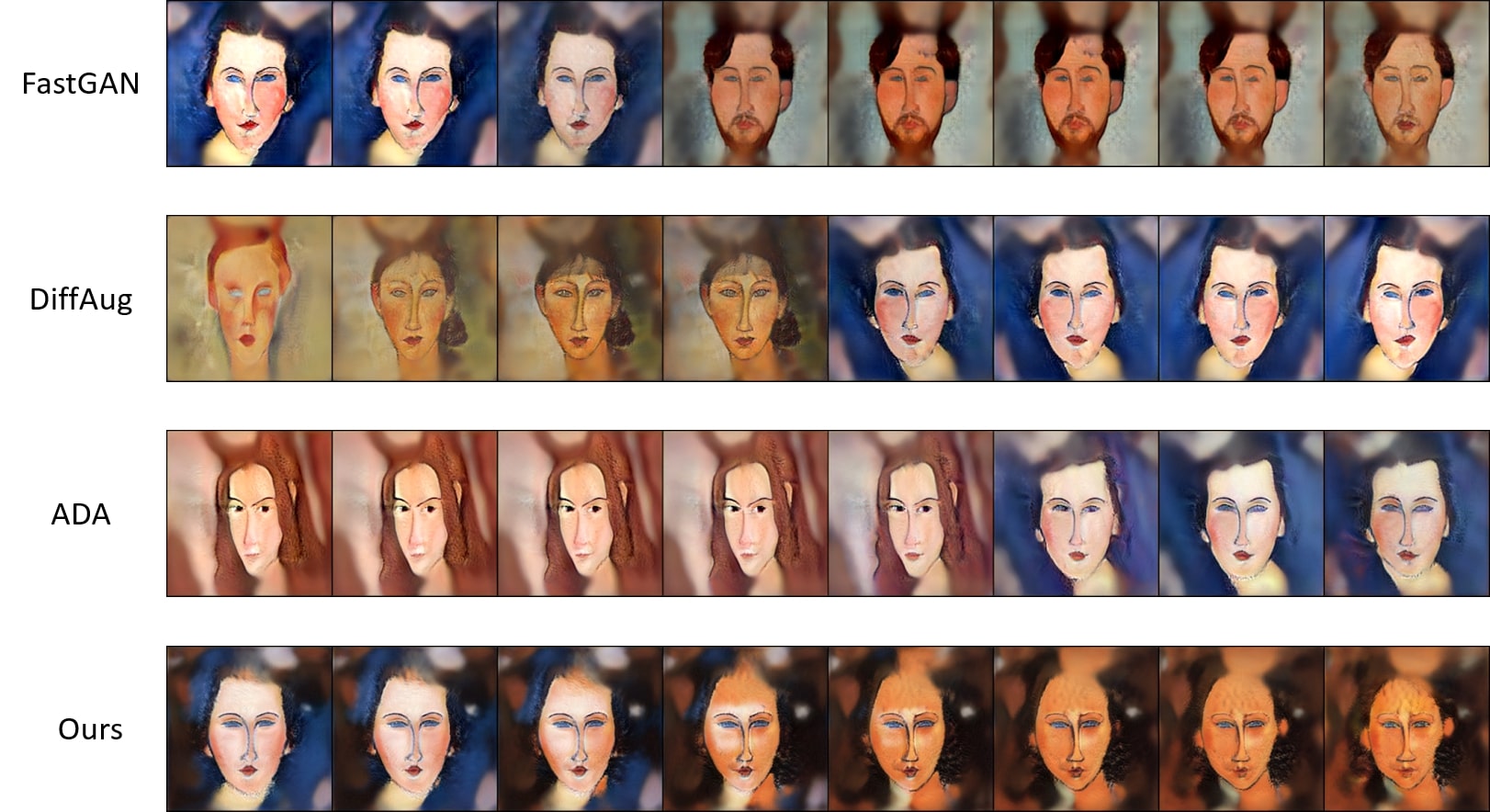}
  \caption{
  Interpolation examples. Baselines clearly display \textit{stairlike} latent transition while ours shows smooth interpolation.
  }
\label{fig:supp_interp2}
\end{figure}

\begin{figure}
\centering
\includegraphics[width=0.95\linewidth]{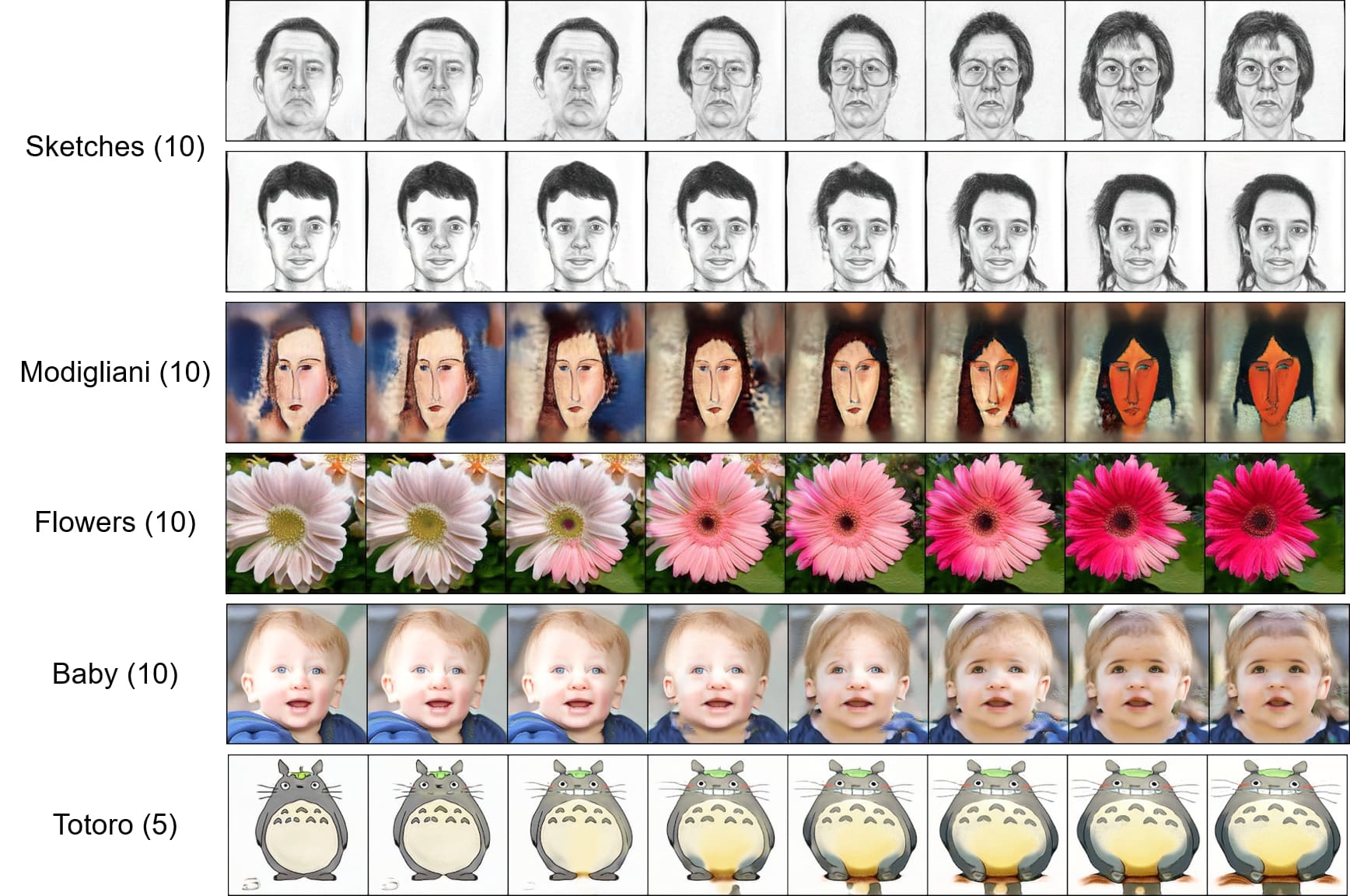}
  \caption{
  More interpolation examples from MixDL. Numbers in the parentheses represent the number of training samples used for each dataset.
  }
\label{fig:supp_interp}
\end{figure}

\begin{figure*}
\includegraphics[width=\linewidth]{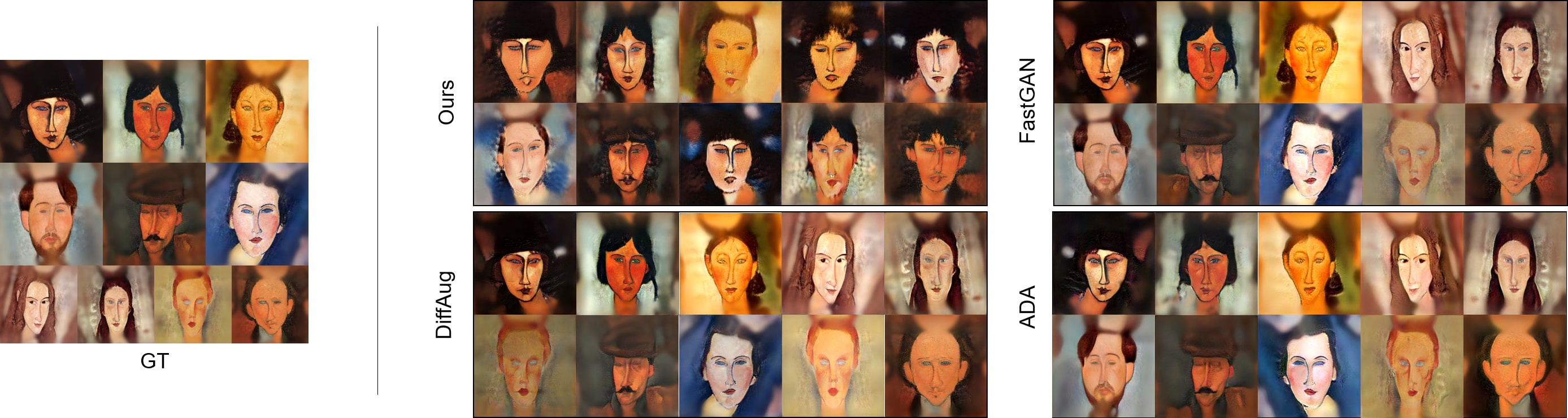}
  \caption{
  Samples from face paintings of Amedeo Modigliani. While the baselines simply replicate the given images, ours produces diverse unseen face images. \textit{Ours} represents samples from StyleGAN2+MixDL.
  }
\label{fig:supp_gen1}
\end{figure*}

\begin{figure}
\centering
\includegraphics[width=0.95\linewidth]{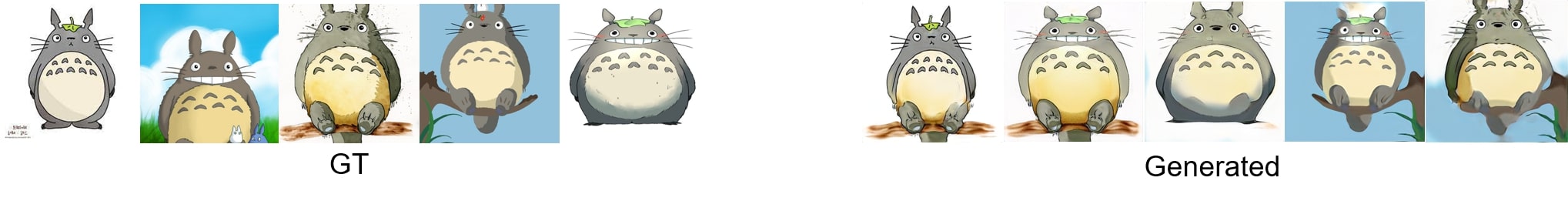}
  \caption{
  MixDL generation result from 5-shot training on Totoro. Although there are only 5 training samples, it combines visual features in a natural way to produce diverse novel samples.
  }
\label{fig:supp_gen3}
\end{figure}

We present latent interpolation results in \cref{fig:supp_interp2} and \cref{fig:supp_interp}. \cref{fig:supp_interp2} shows that MixDL yields smoother latent interpolation compared to baseline methods that show typical \textit{stairlike latent space}. \cref{fig:supp_interp} reaffirms this observation on various datasets. We note that images of Japanese animation character Totoro were crawled from the web, and 5 real samples were used. Additional synthesis results from face paintings of Amedeo Modigliani and illustrations of Totoro are displayed in \cref{fig:supp_gen1} and \cref{fig:supp_gen3}, respectively.


\section{Sample Images from Low-shot Benchmarks}

\begin{figure*}
  \includegraphics[width=0.95\linewidth]{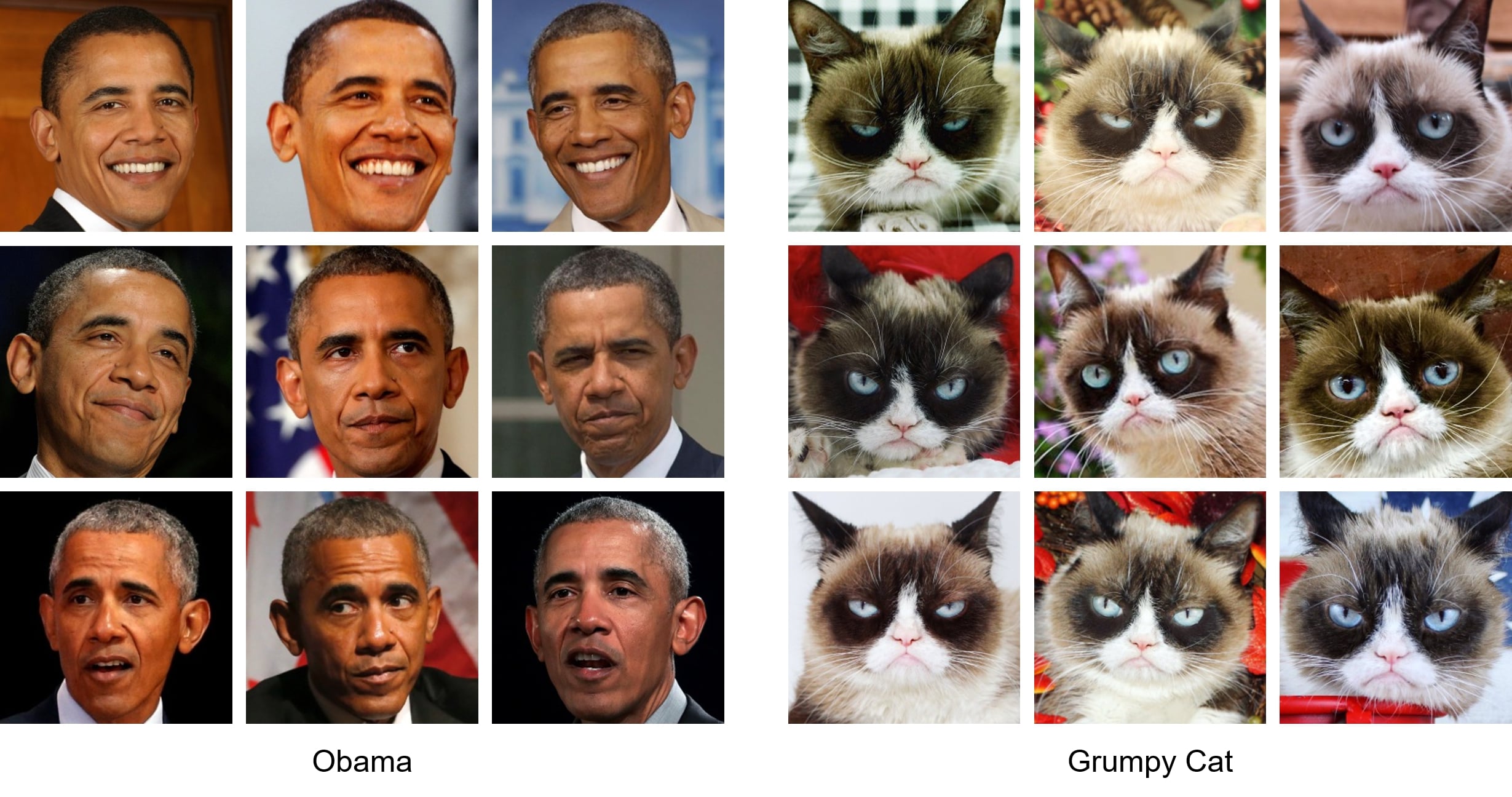}
  \captionsetup{width=\linewidth}
  \caption{
  Random samples from low-shot benchmark datasets, Obama and Grumpy Cat. Since they contain photos of a single character, the intra-diversity is inherently constrained, rendering these benchmarks inappropriate to evaluate generative diversity.
  }
\label{fig:low-shot-sample}
\end{figure*}

In \cref{fig:low-shot-sample}, we present samples from Obama and Grumpy Cat datasets. As they contain images of a single character, the intra-diversity is inherently very limited, which is also demonstrated by the LPIPS measure in Tab. 3 of the main paper.

\section{Naive Application of GAN adaptation}

We display results from naive application of CDC. Since it is very difficult to find a semantically similar source domain for datasets like Pokemon, we naively leverage the source generator trained on FFHQ.
As the source and the target are semantically different, the adaptation does not yield satisfactory outcomes as expected. We can observe the dilemma here as well that in the early iterations, the face shape learned in the source domain is clearly visible while in later stages, the face shape is no longer visible but the model collapses altogether. As CDC preserves distances in the target domain through the correspondence to the source domain, it is not applicable to domains that lack an adequate source dataset to transfer from. MixDL, on the other hand, improves upon CDC in that it enables training generative models with minimal overfitting and mode collapse, without leveraging source domain pretraining. Quantitative evaluations further support the claim as in Tab. 1 of the main paper.

\begin{figure*}
  \includegraphics[width=0.95\linewidth]{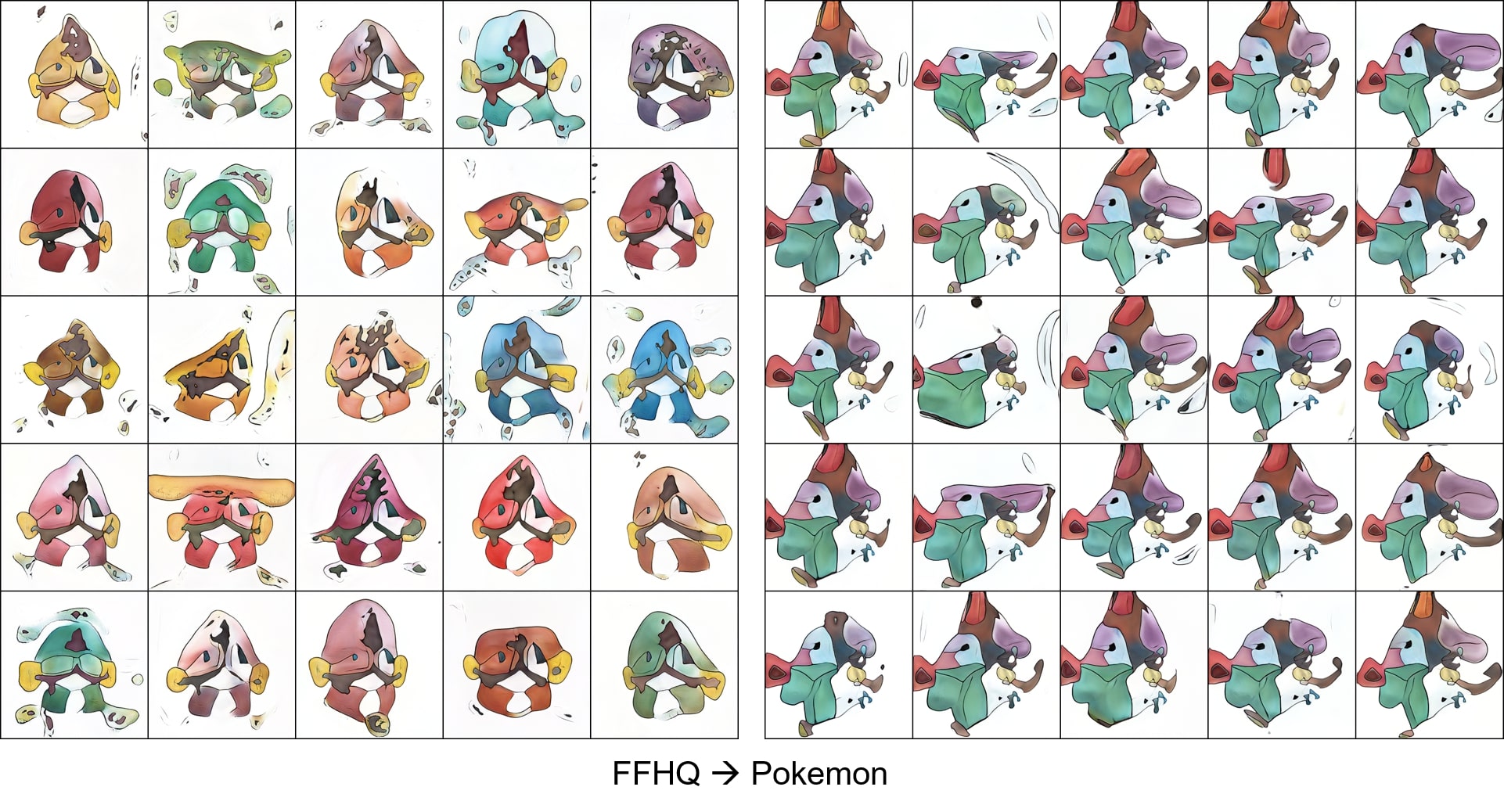}
  \centering
  \captionsetup{width=\linewidth}
  \caption{
  Naive application of CDC from FFHQ to Pokemon. As the authors have pointed out, the adaptation performance degrades when the two domains are semantically different, but it is not straightforward to find a transferable source domain for datasets like Pokemon. We observe clear human face shapes in the early stages \textit{(left)} and mode collapse in later stages \textit{(right)} where the face shape is no longer visible.
  }
\label{fig:CDC1}
\end{figure*}






\clearpage

\bibliographystyle{splncs04}
\bibliography{egbib}
\end{document}